\documentclass[]{TEAI}

\usepackage{helvet}
\usepackage[utf8]{inputenc}
\usepackage[T1]{fontenc}

\usepackage{amsmath,amssymb,mathtools,amsfonts}
\usepackage{graphicx}
\usepackage{subcaption}
\usepackage{wrapfig}
\usepackage{caption}

\usepackage{booktabs}
\usepackage{multirow}
\usepackage{enumitem}
\usepackage{xspace}
\usepackage{tabularx}
\usepackage{nicefrac}

\usepackage{xcolor}
\usepackage{hyperref}
\usepackage{url}

\usepackage{natbib}

\usepackage{catchfile}

\providecommand{\email}[1]{\href{mailto:#1}{#1}}

\newtheorem{definition}{Definition}

\title{SliceGraph: Mapping Process Isomers in Multi-Run Chain-of-Thought Reasoning}

\author{
Kang Chen\textsuperscript{1*},
Junjie Nian\textsuperscript{1*},
Yixin Cao\textsuperscript{1,2$\dagger$},
Yugang Jiang\textsuperscript{1}
}

\affiliation[1]{\mbox{Fudan University}}
\affiliation[2]{\mbox{Shanghai Innovation Institute}}

\correspondence{\email{yxcao@fudan.edu.cn}}
\checkdata[Code]{\url{https://github.com/JunjieNian/SliceGraph}}

\abstract{

}
\begin{document}
\maketitle

\begingroup
\renewcommand{\thefootnote}{}
\footnotetext{* Contributed equally.}
\footnotetext{$\dagger$ Corresponding author.}
\endgroup

\section{Introduction}
\label{sec:intro}

Test-time sampling now routinely draws many CoT trajectories per
problem~\citep{wang2023selfconsistency,yao2023tree,besta2024graph,snell2025ttc},
while self-consistency aggregates them only by final string.  But
the $N$ runs of one problem-model cell are not a bag of independent
samples: they share intermediate computation, fork at decision
points, and rejoin through common subroutines.  Once internal states
are connected by similarity, consistency can be lifted from final
answers to \emph{processes}, and two correct runs are consistent
when they reach overlapping computation.  Most internal analyses so
far remain within-run, studying semantic convergence, step-specific
geometry, or late correctness
signals~\citep{wei2508evolution,sun2604trajectories} rather than
the joint structure across runs.

Building such a graph requires slice-level similarity, for whole-run
is too coarse: two runs can share a long trunk and diverge at one
fork, or reach the same answer by different routes.  Once the graph
is built, the substantive question shifts from selection to geometry.
When many CoTs solve the same problem, do correct solutions converge
to one process route, or do they form several distinct \emph{process
isomers} that share the same final answer through different
intermediate routes?

\begin{wrapfigure}[14]{r}{0.36\linewidth}
  \vspace{-10pt}
  \centering
  \includegraphics[
    width=\linewidth,
  ]{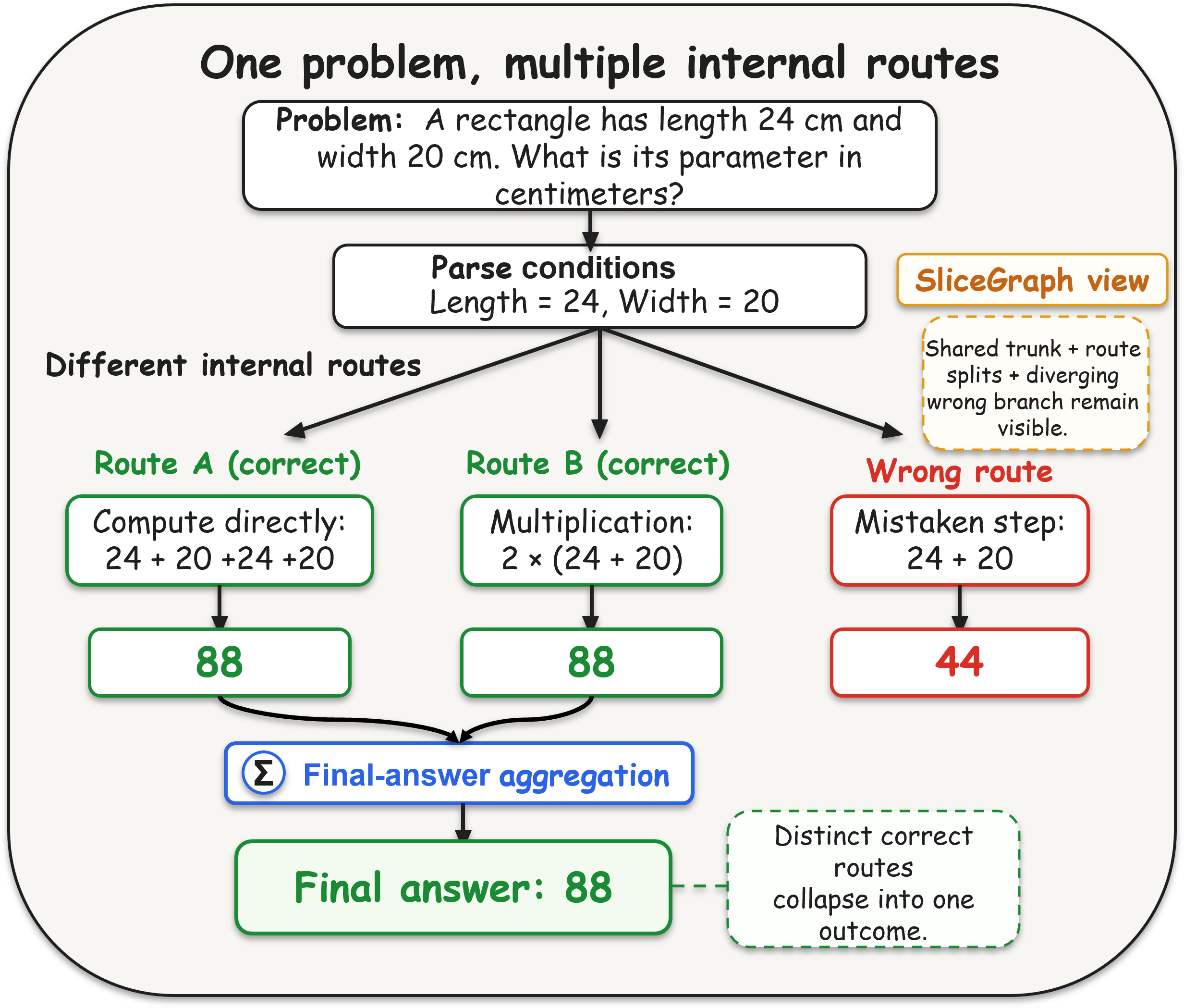}
  \vspace{-8pt}
  \caption{Same-answer runs can follow distinct process families.}
  \label{fig:teaser}
  \vspace{-12pt}
\end{wrapfigure}

Prior graph and topology pipelines on
traces~\citep{xiong2025mapping,zhou2025landscape,tan2025shape}
compress each slice via PCA, UMAP, or sentence transformers before
reading similarities.  In our fixed mutual-$k$NN scaffold, however,
projection- or text-based slice representations distort the route
geometry: PCA compresses correct-only family counts and slice-text
embeddings nearly collapse the atlas
(Section~\ref{sec:representation-validity}).  Unprojected sparse
activation-key Jaccard similarity preserves the multi-route structure
across all three primary models.

We therefore build directly on top-$k$ activation key sets.  Each
slice is a bounded binary key set, and per-problem edges are
mutual-$k$NN under Jaccard distance.  We call this object
\textbf{SliceGraph} and treat it as a measurement target, not a
decoding program.  Biconnected components,
articulation points, and process families are structural units read from
this fixed graph.  Reward diffusion and typed-state transition diagnostics then add
label and temporal information on top of the same scaffold.  Two blinded
annotation studies validate the structural object before any geometric
claim is made, and matched null controls test which claims survive after
graph structure, family labels, answer labels, or temporal order are
perturbed.  Informally, two correct runs that reach the same answer through different
process families are \emph{process isomers}: same problem, same
normalized answer, different route scaffold.  Process isomerism is a route relation, not a
value relation: high-value cores can refine, explain, or separate route
families, but they do not define whether two same-answer traces are
isomers.  High-value cores then show whether those routes traverse shared
or disjoint value basins.

\paragraph{Contributions.}
\begin{itemize}\setlength{\itemsep}{1pt}\setlength{\parskip}{0pt}
  \item \textbf{A process atlas for mapping same-answer route diversity.}  We introduce
    SliceGraph, a post-hoc per-problem-model-cell graph built directly
    from sparse activation-key overlap between CoT slices.  Unlike
    decoding graphs, SliceGraph is a measurement object: it exposes
    shared trunks, forks, partial rejoin points, and route-level
    structure after sampling.
  \item \textbf{Validated structural units.}  We decompose SliceGraph
    into biconnected components, articulation points, process families,
    and label-seeded high-value cores.  Blinded annotation supports BCCs
    as shared reasoning-state units and process families as
    within-family strategy-coherent route units; robustness checks test
    representation choice, graph perturbations, family-label shuffles,
    answer-label shuffles, and temporal-order perturbations.
  \item \textbf{Process isomers behind the same answer.}  On the
    validated atlas, sampled correct CoTs often reach the same
    normalized answer through different correct-only process families.
    We define these family-divergent same-answer traces as process
    isomers.  High-value cores provide a separate value-landscape
    layer, showing that success-associated regions are often
    multi-basin and that route families specialize over those basins.
  \item \textbf{Route dynamics, not just route labels.}  We lift
    block trajectories to a typed-state transition model and show that
    family-specific transition kernels differ significantly under
    matched null controls.
\end{itemize}

\section{Related Work}
\label{sec:related}

\paragraph{Graphs as inference programs versus measurement objects.}
Self-consistency~\citep{wang2023selfconsistency},
Tree-of-Thoughts~\citep{yao2023tree}, and
Graph-of-Thoughts~\citep{besta2024graph} use graphs as decoding
strategies built before or during sampling.  Our graph is instead a
\emph{measurement} object: it is built after sampling from local
activation similarity across many runs, and is used to characterise
process geometry rather than to drive decoding.

\paragraph{Within-run trajectory analysis.}
A line of work analyses LLM reasoning as structured motion in
representation space, including semantic convergence from truncated
trajectories~\citep{wei2508evolution}, step-specific
geometry~\citep{sun2604trajectories}, layer-wise displacement of the
truth signal~\citep{damirchi2026truth}, hidden-state DAG
probes~\citep{zhong2026chains}, and shared decision
pivots~\citep{cho2025pivots}.  These analyses stay within a single
trajectory; we study the joint neighbourhood structure across runs
for one problem.

\paragraph{Graph and topology pipelines.}
The graph and topology pipelines we compare against rely on
intermediate projection: Mapping the
Minds~\citep{xiong2025mapping} clusters CoT paragraphs into NLI
dependency graphs, Landscape of Thoughts~\citep{zhou2025landscape}
projects answer-distance vectors with $t$-SNE, and The Shape of
Reasoning~\citep{tan2025shape} runs persistent homology on step
embeddings.  Activation-set similarity has been used as a
best-of-$N$ selector in Neuron Agreement
Decoding~\citep{chen2025nad}; the same sparse-cache infrastructure
supports temporal explore--exploit phase detection for label-free CoT
scoring~\citep{chen2026nex}. Building on these, we construct the graph directly on
unprojected sparse activation-key Jaccard with no projection or
sentence embedding and then analyse its combinatorial topology;
Section~\ref{sec:representation-validity} shows that PCA and
sentence-embedding pipelines on the same scaffold compress the
multi-route family geometry that SliceGraph captures.

\section{Method}
\label{sec:method}

The method has five layers.  The first three build the route scaffold
on which process isomerism is measured without the label-seeded reward
field; the last two add value-landscape and temporal information on top
of that fixed scaffold.  Graph topology, BCC membership, and
correct-run co-visitation families do not use the reward field; role
labels are interpretive overlays that may use final-answer purity but
do not define process families or process isomers.
Section~\ref{sec:graph-construction} builds the graph from sparse
activation slices.  Section~\ref{sec:bcc} decomposes it into
biconnected components and assigns roles, giving the structural units
later validated in Section~\ref{sec:semantic-validity}.
Section~\ref{sec:families} clusters runs into process families and
introduces the \emph{process isomer} relation that organises the
multi-route geometry of Section~\ref{sec:multi-route-geometry}.
Section~\ref{sec:reward-field-method} defines the label-seeded
reward field and the high-value-core mask, and
Section~\ref{sec:semi-mdp-method} lifts each run to a typed-state
transition model whose family-specific kernels anchor
Section~\ref{sec:family-dynamics}.  All main analyses operate on
non-trivial regions only.

\subsection{SliceGraph construction}
\label{sec:graph-construction}

For each problem-model cell we sample $N{=}64$ separately sampled CoT runs from a
precomputed sparse activation cache.  Each \emph{neuron} is a
(layer, unit) pair indexed by a $32$-bit key, and the active set at
a given token is the top-ranked MLP neurons by the positive part
of the post-gating intermediate (Appendix~\ref{app:cache-details}).  We first aggregate activations into
$32$-token rows, then aggregate every consecutive
$\texttt{sep\_up}$ rows into one coarser reasoning slice; at the
canonical $\texttt{sep\_up}{=}8$, a graph node therefore summarises a
$256$-token segment.  Magnitudes are discarded
after retaining the most-active keys, so each analysed slice is a
binary key set $\mathcal{K}_{r,t}$ with no projection or
text-embedding step.

We connect two slices by their Jaccard overlap,
$d_J(i,j) = 1 - |\mathcal{K}_i \cap \mathcal{K}_j| /
|\mathcal{K}_i \cup \mathcal{K}_j|$, and keep an undirected edge \emph{iff} the relation is mutual in the top-$k$ neighbourhood,
weighting edges by an RBF kernel of $d_J$.  The resulting graph
$\mathcal{G}$ has nodes indexed by run and coarse position, with
edges encoding cross-run local similarity on a single problem.  The
neighbourhood parameter $k$ controls only how many local neighbours
are eligible before the mutuality filter.  Exact cache constants, kernel scale, and
size cap are in Appendix~\ref{app:cache-details}; sensitivity sweeps
are in Appendices~\ref{app:scale-sensitivity}
and~\ref{app:family-construction-sensitivity}.

\begin{figure}[t]
  \centering
  \includegraphics[width=\linewidth]{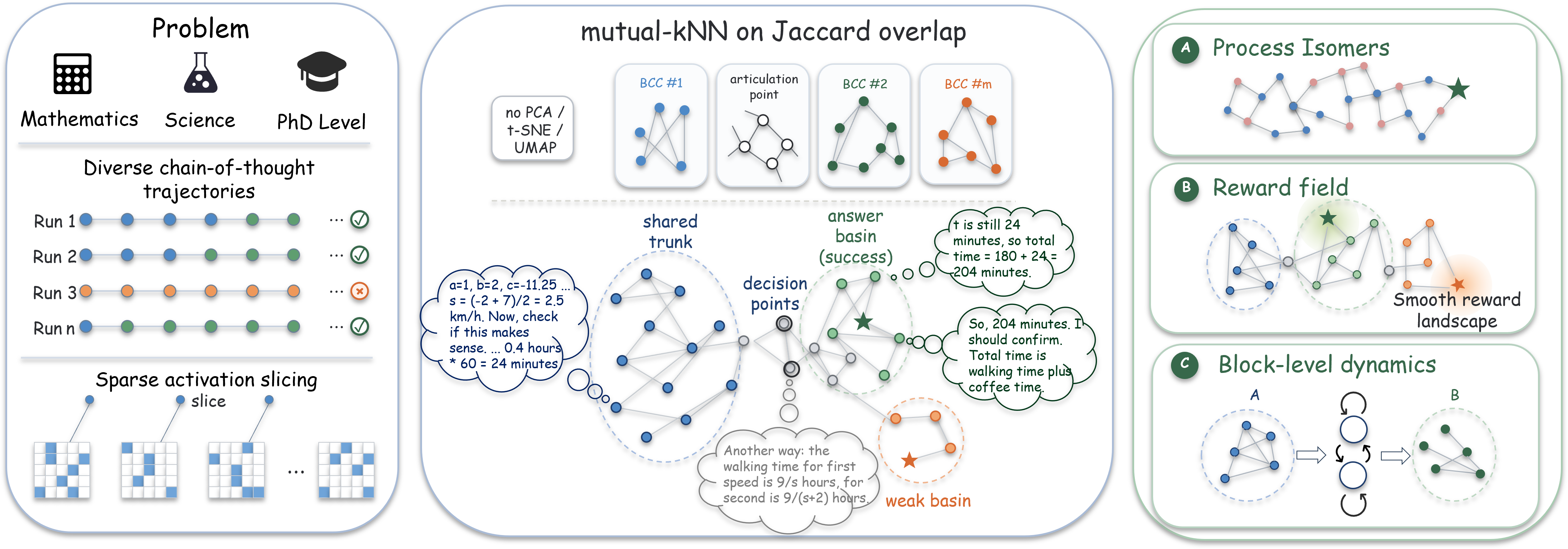}
  \caption{\textbf{SliceGraph pipeline.}
  SliceGraph turns many sampled CoTs for one problem-model cell into an
  activation-key atlas.  Process families are structural units read from
  the fixed graph; label-seeded high-value cores and family-specific
  navigation kernels add answer labels and temporal transitions on top
  of that scaffold.}
  \label{fig:pipeline}
\end{figure}

\subsection{Biconnected-component decomposition and non-trivial regions}
\label{sec:bcc}

We treat $\mathcal{G}$ as a graph of \emph{shared slice-level computation}.  A
\emph{biconnected component} (BCC) is a maximal subgraph that stays
connected after removing any single node, and so groups slices that
are mutually reachable through redundant paths.  An
\emph{articulation point} between two BCCs is a \emph{branch point}
where the reasoning population forks.  We decompose $\mathcal{G}$ component-wise into BCCs and split the
result into trivial $K_2$ \emph{bridges} and non-trivial
$|b|{\ge}3$ blocks.  No largest-connected-component filter is applied:
disconnected components remain eligible, while $K_2$ bridges are
retained as connectors but are not treated as semantic regions.
Sensitivity to this convention is reported in
Appendix~\ref{app:scale-sensitivity}.

We then assign each block one of five roles, namely
\textsc{shared\_trunk}, \textsc{answer\_basin}, \textsc{weak\_basin},
\textsc{decision\_point}, or \textsc{intermediate}, based on its run
coverage, size, dominant-answer purity, and median position.  The
explicit thresholds and bridge-versus-cycle
formulation are in Appendix~\ref{app:method-roles}; role sensitivity
is in Appendix~\ref{app:scale-sensitivity}.  The correspondence between these units and human notions of shared
reasoning state is validated in Section~\ref{sec:semantic-validity};
the same units support the family geometry, label-seeded solution
landscape, and navigation analyses that follow.

\subsection{Process families and process isomers}
\label{sec:families}

Two runs that visit similar sets of regions are likely to share a
route footprint, and may correspond to the same high-level strategy.
We turn this intuition into a clustering by reducing each run to the
set of non-trivial
blocks it visits across all analysed slices,
$\mathcal{S}_r = \bigcup_t \mathcal{C}^{\text{nt}}_{r,t}$,
and forming a weighted Jaccard with rare-block upweighting
$\omega_b = 1/\max(c_b, 0.01)$, where $c_b$ is the fraction of runs
reaching block $b$:
\begin{equation}
  J(r,r') = \frac{\sum_{b \in \mathcal{S}_r \cap \mathcal{S}_{r'}}
                  \omega_b}
                 {\sum_{b \in \mathcal{S}_r \cup \mathcal{S}_{r'}}
                  \omega_b}.
  \label{eq:run-jaccard}
\end{equation}
Repeated visits are discarded, since families are meant to capture
which regions are visited rather than how long a run dwells in
them.  After block-coverage weights are fixed from the full cell, all
headline correct-family statistics construct this run-level co-visitation
graph on normalized-correct runs only, keep edges with $J{\ge}0.05$, and
partition it by Louvain community detection at resolution $1.0$ and
seed $42$.  Incorrect runs are used to define full-cell block coverage,
the reward field, and visual overlays, but they are not nodes in the
correct-only Louvain partition.  Whether
two runs placed in the same family are also judged by humans to use
the same strategy is the second annotation study of
Section~\ref{sec:semantic-validity}.  These families anchor the \emph{process isomer} relation:

\begin{definition}[Process isomer]
\label{def:isomer}
Two correct CoT runs $r,r'$ for the same problem-model cell are
\emph{process isomers} if they reach the same normalized final answer
and are placed in different correct-only process families under the
weighted-Jaccard Louvain partition above.
\end{definition}

\noindent High-value cores introduced next
(Section~\ref{sec:reward-field-method}) serve as a value-landscape
refinement rather than a second isomer criterion.

\subsection{Label-seeded reward field and high-value cores}
\label{sec:reward-field-method}

Families capture which regions a run visits but not which regions
belong to the shared solution landscape.  The reward field is not used
to discover process families and is not treated as an online correctness
predictor; it is a label-seeded explanatory field used only after the
graph and correct-family partition are fixed.  For a block $b$ visited
by runs $\mathcal R_b$, the global seed is a support-shrunk excess
success rate,
\begin{equation}
  v_b^{(0)} =
  \mathbf 1[3\le n_b\le n{-}3]
  \left(\frac{1}{n_b}\sum_{r\in\mathcal R_b} y_r -
  \frac{1}{n}\sum_r y_r\right)
  \!\left(\frac{n_b}{n}\right)^{\!1/2}\!,
  \;\; n_b{=}|\mathcal R_b|.
  \label{eq:reward-seed}
\end{equation}
We then diffuse this seed over a row-normalized block matrix
$\mathbf P$ that combines self-loops, block-cut adjacency, and
binarized symmetrized temporal transitions:
\begin{equation}
  \mathbf{v}^{(t+1)} =
    \alpha\mathbf{v}^{(0)} + (1{-}\alpha)\mathbf{P}\mathbf{v}^{(t)},
  \label{eq:loo-seed}
\end{equation}
with $\alpha{=}0.65$ for $24$ PageRank-style steps.  The run-level
leave-one-out variant, support thresholds, and full adjacency
construction are in Appendix~\ref{app:reward-detail}.  The \emph{high-value
core} $\mathcal{C}$ is the top quartile of strictly-positive field
values on $\mathbf{v}^{(24)}$, and serves as the label-seeded
geometry analysed in Section~\ref{sec:value-landscape}.  When
$\mathcal{C}$ splits into several connected components the cell is
\emph{multi-core}, giving a value-landscape divergence layer
complementary to the family partition.

Section~\ref{sec:value-landscape} uses specialization and coverage
loss as primary alignment readouts, and reports field sharpness as a
secondary diagnostic, each with its natural denominator.
\emph{Family--core footprint specialization} asks whether each
correct-only family concentrates its visited-block footprint on a
subset of high-value-core components.  \emph{Field sharpness} is the
ratio between the maximum of a family-conditioned reward field and the
maximum of the pooled reward field.  \emph{Block-coverage loss}
removes a family and measures the drop in the union of visited
non-trivial blocks; the population readout uses the most
coverage-critical family per cell.  For $K{\ge}2$ high-value-core
components, we write $p_f(k)\propto|B_f\cap C_k|/|C_k|$ for
family~$f$'s normalized core-footprint distribution and report
\begin{equation}
  \mathrm{Spec}=1-\frac{1}{|F_+|}\sum_{f\in F_+}
  \frac{H(p_f)}{\log K},
  \label{eq:family-core-spec}
\end{equation}
where $F_+$ are families with non-zero core footprint; thus $0$ means
diffuse family footprints and $1$ means component-specific footprints.

\noindent\textbf{Core divergence.}\;
For reward-evaluable cells, two correct runs are \emph{core-divergent}
if their non-empty sets of visited high-value-core components are
disjoint.  This relation characterizes value-landscape divergence
rather than defining the primary route partition; it is evaluated only
for runs with at least one high-value-core visit.

\subsection{Family-specific typed-state dynamics}
\label{sec:semi-mdp-method}

To analyse navigation, we lift each run's block trajectory to a
typed-state transition chain
$z_{r,t} = \bigl(\operatorname{role}(b_{r,t}^{\star}),\;
\operatorname{posbin}(\tau_t),\;\operatorname{core}(b_{r,t}^{\star})\bigr)$, where
$b_{r,t}^{\star}$ is the primary non-trivial block at slice $t$ and
$\tau_t\in[0,1]$ is the normalized within-run position binned into
early, mid, and late thirds.  The state alphabet deliberately
includes the high-value-core indicator: dynamics are estimated by
process family, but the states record whether a route is inside or
outside the label-seeded value landscape.  Slices with no non-trivial
block are skipped before compaction; consecutive duplicate typed states
are merged, absorbing correct/wrong terminals are appended, and
Laplace-smoothed family kernels $P_f$ are estimated from compacted
transition counts $C^f_{ij}$:
\begin{equation}
  P_f(i,j)=
  \frac{C^f_{ij}+\alpha_{\mathrm{smooth}}}
       {\sum_{j'} C^f_{ij'}+\alpha_{\mathrm{smooth}}|\mathcal S|},
  \qquad \alpha_{\mathrm{smooth}}{=}0.5 .
  \label{eq:family-kernel}
\end{equation}
The population diagnostic is
row-averaged total variation,
\begin{equation}
  \mathrm{TV}(P_f,P_{f' }) =
  \frac{1}{2|\mathcal S|}\sum_{i\in\mathcal S}
  \|P_f(i,\cdot)-P_{f' }(i,\cdot)\|_1 ,
  \label{eq:family-tv-main}
\end{equation}
where $\mathcal S$ follows the released state alphabet; Appendix~\ref{app:reward-and-dynamics} details the absorbing-terminal convention.
Companion navigation diagnostics are defined in
Appendix~\ref{app:reward-and-dynamics} and are used only for
case-level inspection.

SliceGraph is evaluated as a measurement object for process
geometry and process-isomer structure; using it inside decoding is
left for future work.

\section{Experiments}
\label{sec:experiments}

\subsection{Setup}
\label{sec:setup}

We evaluate SliceGraph on five math and science reasoning benchmarks:
AIME24~\citep{aime24}, AIME25~\citep{aime25}, BRUMO25 and HMMT25 from
MathArena~\citep{balunovic2025matharena}, and GPQA
Diamond~\citep{rein2023gpqa}.  The primary corpus uses three 4B and
8B models sampled with chain-of-thought prompting:
DeepSeek-R1-0528-Qwen3-8B (\textbf{R1-8B}),
Qwen3-4B-Thinking-2507 (\textbf{Think}), and Qwen3-4B-Instruct-2507
(\textbf{Inst}), all run at the canonical configuration.  The primary
corpus contains $60{,}622$ graph-valid analysed trajectories from
$61{,}052$ sampled runs across $954$ problem-model cells, corresponding to
$318$ unique benchmark problems each evaluated under the three
primary models at $N{=}64$ runs per cell.  Run-number sensitivity
suggests that $N{=}64$ retains $95.0\%$ of the process-family
coverage observed on deeper $N{=}256$ caches
(Appendix~\ref{app:scale-sensitivity}).  Subsampling curves further
show that block coverage is near saturation around $N{\approx}32$,
while family and high-value-core counts keep growing as rare-block
neighbourhoods enter the sample.  We run a cross-architecture
replication on DeepSeek-R1-Distill-Llama-8B (\textbf{Llama-8B}), a
within-family cross-scale replication on Qwen3-32B (\textbf{Qwen3-32B}), and a cross-generation
cross-scale replication on the math-specialized
Qwen2.5-Math-72B-Instruct (\textbf{Q2.5-72B}), with per-model results
in Appendices~\ref{app:llama-cross-architecture},
\ref{app:32b-replication}, and~\ref{app:72b-replication}.  Effect sizes
and uncertainty conventions are detailed in
Appendix~\ref{app:stats}.  The reward, multi-core,
typed-state, and family-TV analyses use progressively stricter subsets
carved from the canonical $954$ problem-model cells, with the filtering
chain in Appendix~\ref{app:subset-summary}.  Correctness
and same-answer membership are evaluated after
benchmark-specific answer normalization, and
process-isomer comparisons are made only within the normalized
correct-answer class.

We organise the experiments as a validity ladder: semantic
validation (\S\ref{sec:semantic-validity}), representation
preservation (\S\ref{sec:representation-validity}), family-defined
process-isomer geometry (\S\ref{sec:multi-route-geometry}),
label-seeded value-landscape refinement
(\S\ref{sec:value-landscape}), and value-aware isomer dynamics
(\S\ref{sec:family-dynamics}).

Each RQ is paired with a targeted control: human annotation for
graph-unit semantics, representation ablations for scaffold choice,
pairwise and split-half diagnostics for process isomers,
reward-core alignment readouts for the value landscape, and
family-label/common-support nulls for navigation dynamics
(Table~\ref{tab:claim-ledger} in the appendix).

\subsection{\textbf{RQ1.} Are the graph units semantic?}
\label{sec:semantic-validity}
\label{sec:rq1}

Before reading geometry off the graph, we ask whether its two
discovered units, namely biconnected components and process
families, line up with human notions of shared thinking and shared
strategy.  We answer this with two blinded annotation studies, each
run by two independent annotators on matched within-unit and
across-unit pairs.  Both studies support the semantic interpretation.

\paragraph{Biconnected components correspond to shared reasoning state.}
We constructed $72$ blinded slice-text pairs, half within the same
BCC and half length-matched across BCCs, and asked two independent
annotators to judge whether the two slices represent the same
reasoning operation under matched strict execution-level rubrics.
Annotator~A matches the within/across-BCC labels on $88.9\%$ of pairs
and Annotator~B on $90.3\%$; cross-annotator agreement is
$\kappa{=}\mathbf{0.97}$ on the $n{=}71$ decisive pairs.  The annotation protocol and agreement
statistics are reported in Appendices~\ref{app:annotation-protocol}
and~\ref{app:annotation-bcc}.

\paragraph{Process families are within-family strategy-coherent.}
A separate $77$-pair pack drawn from $8$ problems, with $46$
within-family and $31$ length-matched across-family pairs, asked
whether two runs use the same high-level strategy.  Within-family
pairs reproduce the same-strategy judgement in $\mathbf{94.4\%}$ of
decisive cases; cross-annotator agreement on the across-family slice
is $\kappa{=}\mathbf{0.74}$ with $81.8\%$ overall decisive exact agreement
(Appendix~\ref{app:annotation-family-task}).
Across-family separation is partial at $47.6\%$ different-strategy rate;
families capture finer-grained process differences than ``high-level
strategy,'' so we use \emph{process isomer} at the route-family
granularity.
Task~B should therefore be read as validating within-family coherence
rather than perfect separation of human-named strategies.
Across-family pairs can still share a textbook strategy, because
SliceGraph families are defined at a finer route-footprint granularity;
hence \emph{process isomerism} denotes different graph-measured process
routes, not necessarily different human solution templates.

\subsection{\textbf{RQ2.} Does the activation-key scaffold preserve process geometry?}
\label{sec:representation-validity}
\label{sec:rq2}

If the scaffold distorts process geometry, the isomer and dynamics
claims that follow are built on artefactual structure.  We hold the
mutual-$k$NN framework, BCC decomposition, and family construction
fixed and vary only the slice representation, evaluating each by
two criteria: whether it yields a valid non-trivial BCC scaffold,
and whether the resulting family geometry preserves the canonical
multi-route multiplicity.

\paragraph{Evidence.}
Under the fixed scaffold, activation-key Jaccard preserves the
highest and most stable correct-family multiplicity across primary
models (Table~\ref{tab:repr-ablation-main}; per-model breakdown in
Appendix~\ref{app:repr-ablation}).  Cosine and overlap on the same
sparse keys are close, supporting the activation-key object rather
than a single metric artefact.  PCA-$50$ remains graph-valid but
compresses family counts by ${\sim}1$ per cell, while slice-level
MiniLM nearly collapses the atlas: only $21/954$ cells yield a valid
decomposition, with at most $11$ valid cells for any primary model.

\begin{table}[ht]
\caption{Representation ablation under a fixed mutual-$k$NN scaffold.
Multi-fam rate\,/\,mean correct families per valid cell.
Canonical rows evaluate all $954$ cells; collapsed representations
report their valid-cell counts (see Appendix~\ref{app:repr-ablation}).}
\label{tab:repr-ablation-main}
\centering\footnotesize
\begin{tabular}{@{}lccc@{}}
\toprule
Scaffold representation & R1-8B & Think & Inst \\
\midrule
\textbf{Activation keys, Jaccard (ours)} & $\mathbf{89.3/5.14}$ & $\mathbf{86.2/4.61}$ & $\mathbf{81.1/4.21}$ \\
Activation keys, cosine   & $88.1/4.80$ & $85.5/4.36$ & $79.6/3.87$ \\
PCA-$50$ + cosine          & $81.4/3.38$ & $81.4/3.33$ & $73.8/3.00$ \\
MiniLM slice-text          & \multicolumn{3}{c}{\emph{collapsed} ($21/954$ valid cells; ${\le}11$ per model)} \\
\bottomrule
\end{tabular}
\end{table}

PCA preserves an activation-based object but compresses set-overlap
contrast; MiniLM collapses because same-problem topical similarity
saturates execution-level contrast.  The unprojected activation-key
scaffold therefore occupies the narrow band between collapse and
over-fragmentation that preserves the route structure validated in
RQ1, and we use Jaccard as the canonical metric.

\subsection{\textbf{RQ3.} Are same-answer correct CoTs route-isomeric?}
\label{sec:multi-route-geometry}
\label{sec:rq3-multi-route}
\label{sec:rq3}
\label{sec:heterogeneity}

\begin{figure}[t]
  \centering
  \begin{subfigure}[t]{0.32\linewidth}
    \centering
    \includegraphics[width=\linewidth]{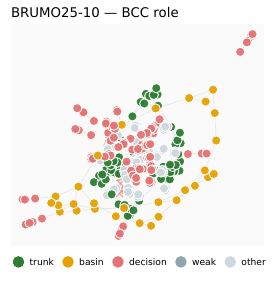}
  \end{subfigure}\hfill
  \begin{subfigure}[t]{0.32\linewidth}
    \centering
    \includegraphics[width=\linewidth]{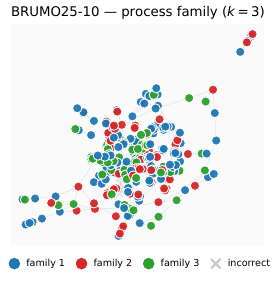}
  \end{subfigure}\hfill
  \begin{subfigure}[t]{0.32\linewidth}
    \centering
    \includegraphics[width=\linewidth]{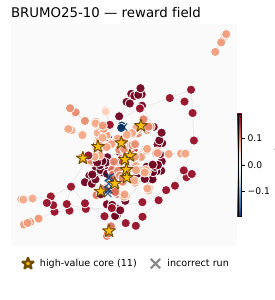}
  \end{subfigure}
  \caption{Three readouts of a single SliceGraph for BRUMO25-10
  (Inst, $64$ runs, $60$ correct).
  Every node is one CoT \textbf{slice}; $273$ slices in non-trivial
  biconnected components are shown, with trivial $K_2$ bridge components omitted.
  \textbf{Left:} BCC block-type role (trunk, basin, decision, weak, other).
  \textbf{Middle:} correct-only process families ($k{=}3$ families;
  $\times$\,=\,incorrect runs).
  \textbf{Right:} propagated reward field seeded by per-block correctness
  deviation; $\bigstar$ marks the centroid of each high-value-core
  component ($11$ cores).
  Note that family colours and core locations split differently,
  illustrating that route families and value cores are related but
  non-identical layers.}
  \label{fig:case-overlays}
\end{figure}

\begin{figure}[t]
  \centering
  \begin{subfigure}[t]{0.32\linewidth}
    \centering
    \includegraphics[width=\linewidth]{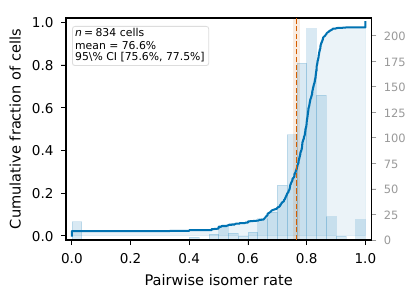}
    \caption{Isomer rate ECDF.}
  \end{subfigure}\hfill
  \begin{subfigure}[t]{0.32\linewidth}
    \centering
    \includegraphics[width=\linewidth]{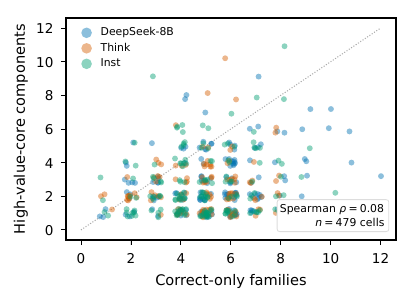}
    \caption{Families vs.\ cores.}
  \end{subfigure}\hfill
  \begin{subfigure}[t]{0.32\linewidth}
    \centering
    \includegraphics[width=\linewidth]{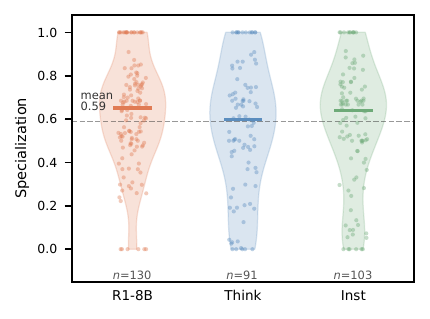}
    \caption{Core specialization.}
  \end{subfigure}
  \caption{\textbf{RQ3/RQ4 population geometry.}
  (a)~ECDF of pairwise family-isomer rates over $834$ cells;
  vertical lines mark the bootstrap $95\%$~CI.
  (b)~Correct-family count vs.\ core components on $479$
  reward-evaluable cells (Spearman $\rho{=}0.08$): the two layers are
  non-redundant in count.
  (c)~Family--core specialization on $324$ multi-core cells
  (bars\,=\,medians): despite weak count correlation, families
  spatially specialize over distinct core footprints
  (pooled mean\,$= 0.59$).}
  \label{fig:rq3-isomer-geometry}
\end{figure}

Process isomerism is a family-defined relation
(Definition~\ref{def:isomer}): two correct runs are isomers when they
reach the same normalized answer but belong to different correct-only
process families.  We first establish how prevalent this relation is
using only the family partition; the reward field and typed-state
dynamics are layered on in
\S\ref{sec:value-landscape}--\S\ref{sec:dynamics}.
Figure~\ref{fig:case-overlays} visualises three layers of one atlas;
Figure~\ref{fig:rq3-isomer-geometry} gives population diagnostics and
Table~\ref{tab:model-dataset-breakdown} per-model breakdowns.

\paragraph{Route multiplicity.}
Before introducing the reward field, $\mathbf{85.5\%}$ of the $954$
canonical cells contain more than one correct-only process family.
At the pair level, $\mathbf{76.6\%}$ of normalized-correct run pairs
are family-isomeric on average, with a problem-clustered bootstrap
$95\%$ CI of $[75.6,77.5]\%$
(Figure~\ref{fig:rq3-isomer-geometry}a).

\paragraph{Sampling robustness.}
Split-half resampling preserves the signal on a stricter
split-valid subset: both halves are multi-family in $97.2\%$ of cells
($n{=}779$).  A held-out atlas
built from $32$ train runs and projected onto $32$ test runs retains
$91\%$ multi-family cells; fixing the correct-run count to $m{=}4$
still yields $58\%$ isomer rate
(Appendices~\ref{app:heldout-atlas},
\ref{app:controlled-isomer-rate}).
A resolution$\times$threshold sweep confirms stability
(Appendix~\ref{app:family-construction-sensitivity}).
Low-isomer cells remain, and SliceGraph recovers multi-route
structure when present rather than forcing it.
The signal is not driven by bridge-like length fragmentation: including
$K_2$ bridges barely changes the multi-family rate
($85.5\%{\to}86.8\%$) but overfragments the partition, inflating mean
families from $4.66$ to $11.27$ and raising the between-family
length-variance ratio from $0.46$ to $0.70$
(Appendix~\ref{app:scale-sensitivity}).
Thus, the same correct answer is usually not a single route endpoint but a set
of graph-separated route footprints.

\begin{table*}[t]
\caption{Dataset-level route multiplicity for the primary corpus.
Entries: \% cells with ${\ge}2$ correct-only families\,/\,mean correct families.
Full six-model breakdown including replication models in
Table~\ref{tab:model-dataset-full}.}
\label{tab:model-dataset-breakdown}
\label{fig:specialization}
\label{fig:isomer-evidence}
\centering\footnotesize
\begin{tabular}{@{}lcccccc@{}}
\toprule
Model & AIME24 & AIME25 & BRUMO25 & GPQA & HMMT25 & All \\
\midrule
R1-8B  & 86.7\,/\,4.87 & 86.7\,/\,4.80 & 86.7\,/\,4.57 & 92.4\,/\,5.50 & 76.7\,/\,3.97 & 89.3\,/\,5.14 \\
Think        & 93.3\,/\,5.07 & 86.7\,/\,5.10 & 96.7\,/\,4.77 & 85.9\,/\,4.60 & 70.0\,/\,3.57 & 86.2\,/\,4.61 \\
Inst         & 83.3\,/\,4.20 & 70.0\,/\,3.43 & 76.7\,/\,3.70 & 87.9\,/\,4.72 & 50.0\,/\,2.17 & 81.1\,/\,4.21 \\
\bottomrule
\end{tabular}
\end{table*}

\subsection{\textbf{RQ4.} How do route isomers align with the label-seeded value landscape?}
\label{sec:value-landscape}
\label{sec:rq4-multi-route}
\label{sec:rq4}
\label{sec:reward-field}

We now add the reward field (\S\ref{sec:reward-field-method}) as a value
contour over the route atlas.  With the correct-only family partition
fixed, the field marks where correctness-associated mass concentrates
after label-seeded diffusion.  The question shifts from route
multiplicity to route--value alignment: do same-answer families share one
high-value component, or specialize over multiple?
Core counts are treated as descriptive geometry; label-permutation
stress tests are in Appendix~\ref{app:dynamics-graph-null}.

\paragraph{Evidence.}

\begin{table}[ht]
\caption{Value-landscape readouts for RQ4.  The first three rows test
non-redundancy between route families and high-value cores; the last
three test spatial alignment and coverage.}
\label{tab:rq4-landscape}
\centering\footnotesize
\begin{tabular}{@{}lll@{}}
\toprule
Readout & Value & Subset \\
\midrule
Multi-core rate & $\mathbf{67.6\%}$ & $479$ reward-evaluable \\
2nd\,/\,1st core size & $0.54$ & $324$ multi-core \\
Family--core count $\rho$ & $0.08$ & $479$ reward-evaluable \\
\midrule
Specialization & $\mathbf{0.59}$ & $324$ multi-core \\
Coverage loss $>10\%$ & $\mathbf{94.5\%}$ & $816$ multi-family \\
Core-divergence uplift & $+1.3$\,pp & $834$ isomer-eligible \\
\bottomrule
\end{tabular}
\end{table}

Of $479$ reward-evaluable cells, $324$ are multi-core
(Table~\ref{tab:rq4-landscape}), with mean second-to-first core
size ratio 0.54. This value granularity is not a relabelling of route
multiplicity, since family and core component counts correlate weakly
(Figure~\ref{fig:rq3-isomer-geometry}b).  Yet they align spatially
(Figure~\ref{fig:rq3-isomer-geometry}c)---families specialize over
distinct core footprints and are individually coverage-critical---while
a disjoint-core criterion adds only $1.3$\,pp beyond the family-isomer
relation.  High-value cores therefore refine the
route map without redefining process isomerism.

\subsection{\textbf{RQ5.} Do process isomers navigate the atlas differently?}
\label{sec:family-dynamics}
\label{sec:rq4-dynamics}
\label{sec:rq4-process-dynamics}
\label{sec:rq5-dynamics}
\label{sec:rq5}
\label{sec:dynamics}

\begin{figure*}[t]
\centering
\begin{subfigure}[t]{0.29\textwidth}
  \centering
  \includegraphics[width=\linewidth]{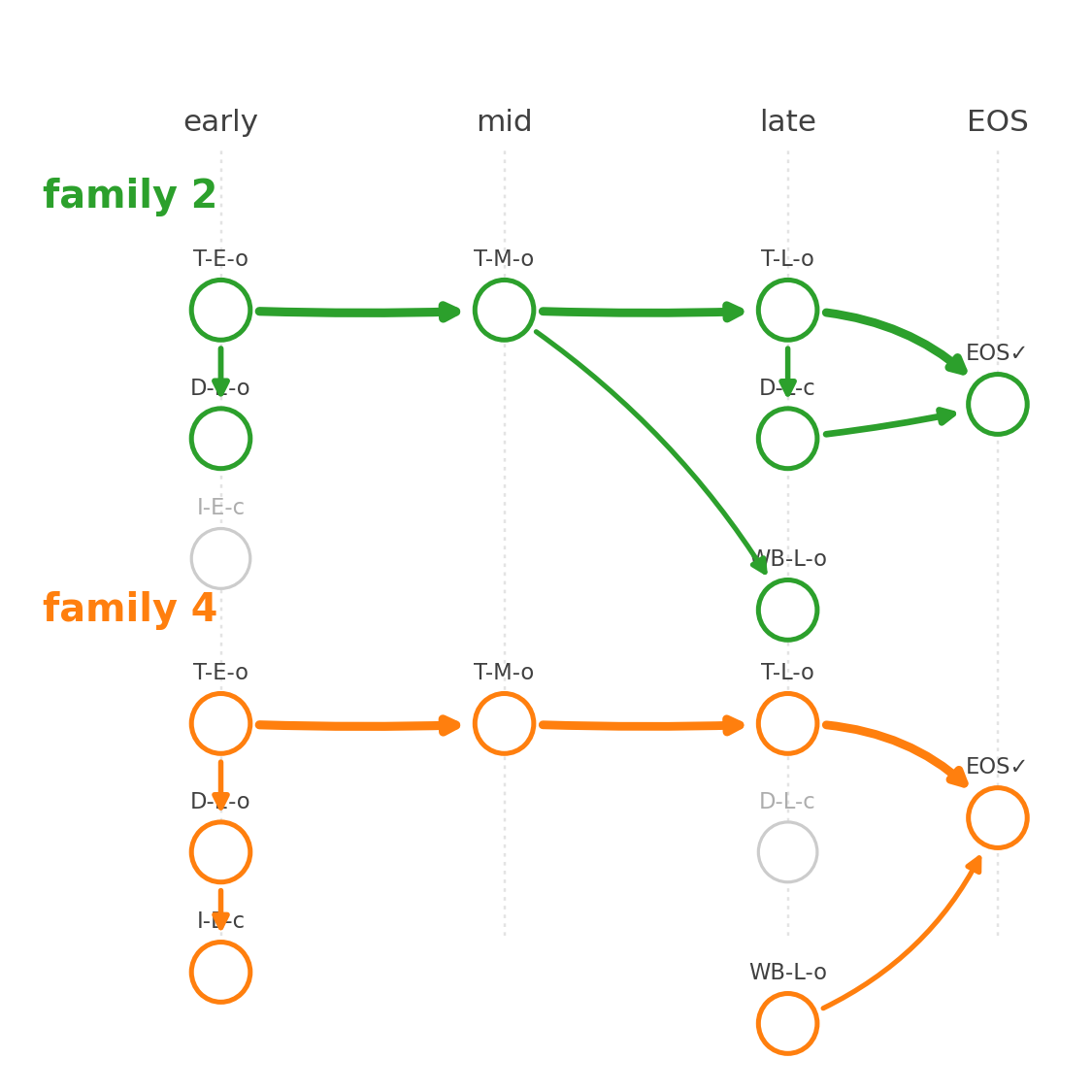}
\end{subfigure}\hfill
\begin{subfigure}[t]{0.29\textwidth}
  \centering
  \includegraphics[width=\linewidth]{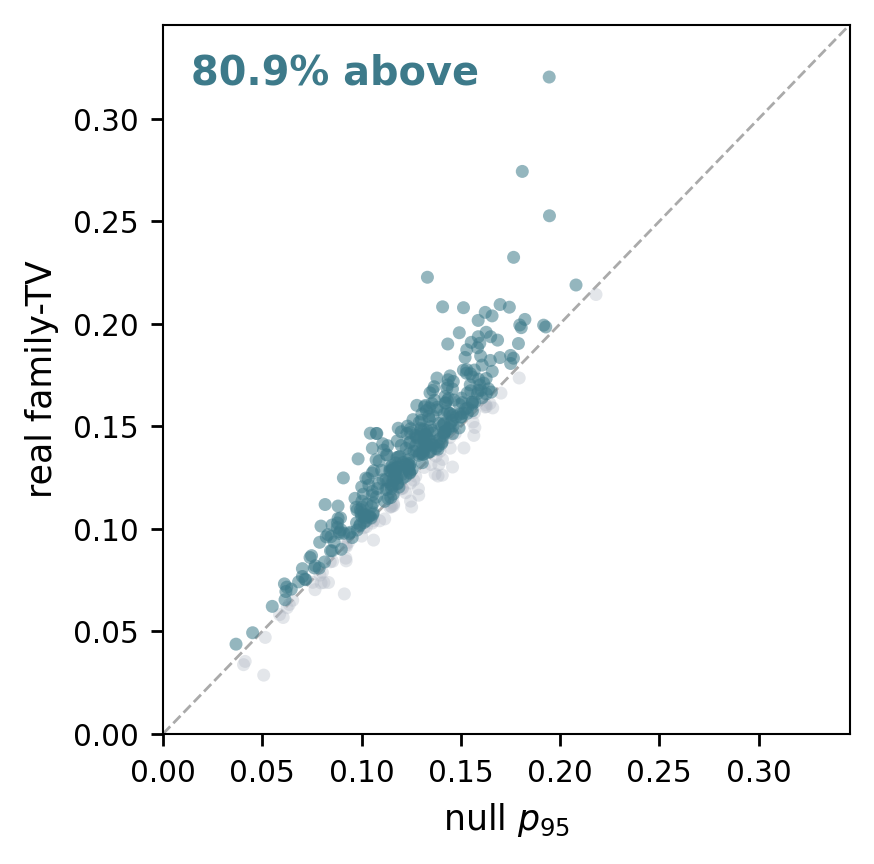}
\end{subfigure}\hfill
\begin{subfigure}[t]{0.29\textwidth}
  \centering
  \includegraphics[width=\linewidth]{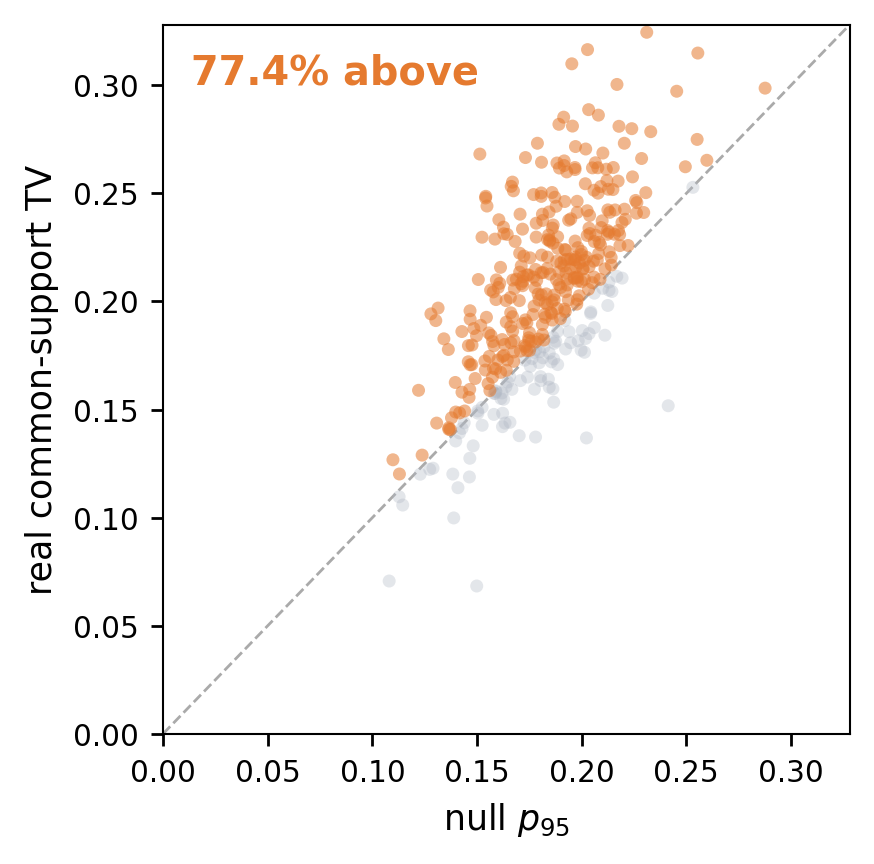}
\end{subfigure}

\caption{\textbf{Family-specific typed-state dynamics.}
(a)~Two families share a typed-state vocabulary but place different
transition mass.
(b,\,c)~Real family-TV and common-support TV versus size-preserving
label-shuffle $p_{95}$; points above the diagonal exceed the matched
null.}
\label{fig:dynamics-schematic}
\end{figure*}

The value layer shows where route families concentrate; it does not yet say
whether they traverse the atlas differently.  We turn the same atlas into a
typed-state traffic model (\S\ref{sec:semi-mdp-method}), estimating one
transition kernel per family and comparing between-family TV to a
size-preserving label-shuffle null.  Common-support TV restricts rows to
states visited by both families, ruling out disjoint support;
Figure~\ref{fig:dynamics-schematic} shows both diagnostics.

\paragraph{Family-specific kernels carry route-specific signal.}
On the $403$-cell core-eligible multi-family subset, row-averaged TV
exceeds the label-shuffle $p_{95}$ in $\mathbf{80.9\%}$ of cells
(median $z{=}3.14$). The result is not just a
static-footprint effect: family-TV is weakly correlated with footprint
specialization (Pearson $-0.07$). Nor is it explained only by disjoint support: common-support TV preserves the
direction: $\mathbf{77.4\%}$ of cells exceed matched $p_{95}$, with
median $z{=}3.45$ and median $7$ shared typed states per family pair.
Thus families differ not only in where they concentrate but also in how
they move through the typed-state atlas.
Appendix~\ref{app:dynamics-graph-null},~\ref{app:family-tv-null} give stress tests and null details;
.

\paragraph{Cross-model replication across RQ3--RQ5.}
Llama-8B tests architecture transfer, Qwen3-32B within-family scaling,
and Q2.5-72B saturation.  Llama-8B and Qwen3-32B match the
route-family, value-landscape, and family-TV pattern; Q2.5-72B
attenuates multiplicity but retains the qualitative geometry on
unsaturated GPQA
(Appendices~\ref{app:llama-cross-architecture},
\ref{app:32b-replication}, \ref{app:72b-replication}).

\section{Conclusion}
\label{sec:conclusion}

SliceGraph is a post-hoc measurement atlas, not a decoder: it maps sparse
neuron-activation keys from multi-run CoTs into a graph of shared and diverging
process routes. On this activation-key atlas, same-answer correct CoTs are often
process-isomeric. High-value cores and family-specific kernels further show how
route families align with value basins and differ in navigation dynamics. Together,
validated graph units, activation-key scaffolds, reward-field-free isomer statistics,
value-layer refinement, and null-controlled dynamics reveal non-monolithic
same-answer reasoning.

\section{Limitations and broader impact}
\label{sec:discussion}
\label{sec:limitations}
\label{sec:robustness}

SliceGraph inherits the limits of observational activation analysis: it requires
hooks and cached sparse activations, depends on slice/block conventions, and may
under-sample rare routes because annotations and long-tail transitions are finite. Its label-seeded reward field is descriptive,
so label permutations are stress tests of that layer rather than one-sided nulls
for multi-core counts (Appendix~\ref{app:dynamics-graph-null}). Claims are scoped to
non-code math/science; code-generation tasks may differ. Broader impacts are
diagnostic: SliceGraph can audit route diversity, collapse, and over-concentration,
but selection or early-stopping use could amplify sampling biases or suppress
rare correct routes.

\clearpage
\bibliographystyle{plainnat}
\bibliography{references}

\appendix

\section*{Appendix roadmap}
\begin{itemize}\setlength{\itemsep}{1pt}
\item \textbf{Implementation and count conventions}:
      analysis subsets~(\S\ref{app:subset-summary});
      cache pipeline and graph hyperparameters~(\S\ref{app:cache-details});
      block roles and bridge-vs-cycle convention~(\S\ref{app:method-roles});
      reproducibility and statistics~(\S\ref{app:details-trailer}).
\item \textbf{Representation and family robustness}:
      block-scale and hyperparameter sensitivity~(\S\ref{app:scale-sensitivity});
      representation ablation per pipeline$\times$model~(\S\ref{app:repr-ablation});
      fixed-scaffold split-half stability~(\S\ref{app:split-half});
      held-out atlas validation~(\S\ref{app:heldout-atlas});
      correct-run controlled isomer rate~(\S\ref{app:controlled-isomer-rate});
      family-construction sensitivity~(\S\ref{app:family-construction-sensitivity});
      isomer-discovery curves under run subsampling~(\S\ref{app:isomer-discovery-curves}).
\item \textbf{Process-isomer geometry and dynamics}:
      reward-field construction~(\S\ref{app:reward-detail});
      multi-core / multi-family population summary~(\S\ref{app:reward-and-dynamics});
      per-model block-type localisation~(\S\ref{app:reward-field-per-model});
      family-specific navigation diagnostics~(\S\ref{app:reward-dynamics-table});
      typed-state sensitivity and scope~(\S\ref{app:dynamics-alpha-note});
      null-control ledger and label-permutation reward stress test~(\S\ref{app:dynamics-graph-null});
      family-TV null construction~(\S\ref{app:family-tv-null});
      worked route report on GPQA-71~(\S\ref{app:route-report}).
\item \textbf{Human annotation protocol}:
      family detection robustness~(\S\ref{app:family-detection-robust});
      annotation protocol~(\S\ref{app:annotation-protocol});
      Task~A (BCC shared-state)~(\S\ref{app:annotation-bcc});
      Task~B (process families)~(\S\ref{app:annotation-family-task});
      qualitative annotation cases~(\S\ref{app:qual-cases}).
\item \textbf{Cross-model replications}:
      Q2.5-72B cross-generation cross-scale~(\S\ref{app:72b-replication});
      Qwen3-32B within-family cross-scale~(\S\ref{app:32b-replication});
      Llama-8B cross-architecture~(\S\ref{app:llama-cross-architecture}).
\end{itemize}

\vspace{0.5em}

\section{Additional Experimental Details}
\label{app:details}

\subsection{Analysis Subsets and Count Conventions}
\label{app:subset-summary}

\emph{Unique runs} are sampled trajectories before graph construction;
\emph{analysed rows} are runs surviving slice selection.  The second
count is slightly smaller where individual runs yield no valid slice
path.

\paragraph{Count ledger.}
The canonical primary atlas corpus has $318$ unique benchmark
problems evaluated under three primary models, giving
$318\times 3 = 954$ problem-model cells.  At $N{=}64$ generations per
cell this yields $61{,}056$ sampled generations; after dropping $4$
incomplete cache records, the corpus contains $61{,}052$ runs.
Graph-valid filtering further drops $430$ runs whose stratified slice
sampling yields an empty path, leaving $60{,}622$ analysed rows.  No
largest-connected-component filter is applied: BCCs are computed
component-wise on the selected graph, so disconnected rare routes are
not silently removed.  We reuse ``$954$-cell corpus'' as shorthand
for these $954$ problem-model cells.  The reward-evaluable subset is
stricter: after graph-valid filtering a cell must have at least $3$
correct and $3$ non-correct runs, at least two non-trivial blocks with
non-empty run--block sequences, at least one seed-eligible block
($3\le n_b\le n-3$ visitors), and non-empty strictly positive support in
$\mathbf v^{(24)}$.  In the exported $479$ reward-evaluable cells the
minimum observed counts are $3$ correct runs, $3$ non-correct runs,
$4$ non-trivial blocks, $3$ seed-eligible blocks, and one positive
high-value-core block.

\begin{table}[h]
\caption{Main analysis subsets used throughout the atlas version of
the paper.}
\label{tab:appendix-subsets}
\centering\footnotesize
\begin{tabular}{@{}p{0.32\linewidth}p{0.14\linewidth}p{0.16\linewidth}p{0.30\linewidth}@{}}
\toprule
Analysis & Cells & Runs / rows & Notes \\
\midrule
Canonical primary corpus &
954 & 60{,}622 analysed rows &
3 primary models, 5 non-code datasets, canonical
$\texttt{sep\_up}{=}8$, $k{=}6$, mutual-$k$NN \\

Reward-evaluable solution landscape &
479 problem-model cells & cell-level subset &
Requires $\ge 3$ correct and $\ge 3$ non-correct runs, $\ge 2$
non-trivial blocks, at least one seed-eligible block, and non-empty
positive field support \\

Core-eligible multi-family subset &
403 problem-model cells & family-level kernels &
Multi-family cells with at least one high-value core; used for
family-TV and joint family/core checks; alignment readouts use
natural denominators in Table~\ref{tab:multi-modal} \\

Cross-architecture replication &
258 & 16{,}512 analysed rows &
DeepSeek-R1-Distill-Llama-8B on AIME24, AIME25, and GPQA \\

Within-family cross-scale replication &
318 & 20{,}352 analysed rows &
Qwen3-32B (thinking mode) on the full 5-dataset non-code suite \\

Cross-generation cross-scale replication &
318 & 20{,}196 analysed rows &
Qwen2.5-Math-72B-Instruct on the full 5-dataset non-code suite \\

Process-family semantics (annotation) &
8 cases & 102 runs &
$46$ within-family + $31$ across-family judged pairs; two
independent non-author annotators \\

BCC reasoning-state semantics (annotation) &
36 cells & 72 slice pairs &
$36$ within-BCC + $36$ matched across-BCC pairs; two independent
non-author annotators \\
\bottomrule
\end{tabular}
\end{table}

\paragraph{Derivation chain.}
The main atlas subsets form the strict refinement chain
\[
954 \;\to\; 479 \;\to\; 403,
\]
namely canonical corpus $\to$ reward-evaluable cells $\to$
core-eligible multi-family cells.  The alignment readouts in
Table~\ref{tab:multi-modal} then use natural denominators: for example,
specialization is averaged on the $324$ multi-core reward-evaluable
cells, while family-TV uses the $403$-cell subset.

\subsection{Activation-Cache and Graph Implementation}
\label{app:cache-details}

\paragraph{Sparse activation logging.}
We record post-gating MLP intermediates under the standard
SwiGLU/GeGLU convention used by Qwen3, DeepSeek-R1-Distill-Qwen,
DeepSeek-R1-Distill-Llama, and Qwen2.5-Math.  Let
$u_{\ell,t},g_{\ell,t}\in\mathbb{R}^{d_{\text{ff}}}$ denote the FFN
``up'' and ``gate'' pre-activations at layer $\ell$ and token $t$,
and let $\phi(\cdot)$ denote the gating nonlinearity (SiLU here).  We
score each neuron by the positive part of its post-gating
contribution,
\begin{equation}
  h_{\ell,t} \triangleq \phi(g_{\ell,t}) \odot u_{\ell,t},
  \qquad
  a_{k,t} \triangleq \max\bigl(0,(h_{\ell,t})_j\bigr),
  \label{eq:cache-act}
\end{equation}
with each neuron keyed across layers as
\begin{equation}
  k = (\ell \ll 16) \; \big| \; j.
  \label{eq:cache-key}
\end{equation}
At each token we retain the top \texttt{global\_topk}$=2000$ neurons
by $a_{k,t}$.

\paragraph{Window aggregation and slice keys.}
Tokens are grouped into rows of \texttt{slice\_size}$=32$.  For row
$r$ with token-index set $T_r$, we form
$A_{k,r}=\sum_{t\in T_r} a_{k,t}$ and keep the
\texttt{slice\_topk}$=500$ neurons with largest mass:
\begin{equation}
  \mathcal{K}_r =
  \mathrm{top}_{K{=}500}\{k : A_{k,r}\},
  \qquad |\mathcal{K}_r|\le K.
  \label{eq:cache-keys}
\end{equation}
The analysis pipeline then aggregates every consecutive
$\texttt{sep\_up}{=}8$ rows into one analysed slice, so one analysed
slice summarises $256$ tokens of CoT.

\paragraph{Graph construction.}
Canonical SliceGraph construction uses mutual-$k$NN with $k{=}6$,
distance scale $\sigma{=}0.35$, and size cap $M{=}2{,}600$ slices per
problem.  Problems exceeding $M$ are stratified-subsampled over runs
and positions before graph construction.  Edge weights are RBF
transforms of Jaccard distance on the sparse key sets.

\subsection{Block Roles, Thresholds, and Bridge-vs-Cycle Convention}
\label{app:method-roles}

The five block roles used in \S\ref{sec:bcc} are assigned from each
block's run coverage $c_b$, size $|b|$, dominant-answer purity
$\pi_b$, unique run count $n_{\text{runs}}(b)$, and median position
$\operatorname{medpos}(b)$:
\begin{equation}
  \text{role}(b) = \begin{cases}
    \textsc{shared\_trunk}
      & c_b > 0.4,\; |b| \ge 6,\;
        \operatorname{medpos}(b) \le q_{0.5}^{\text{trunk}} \\
    \textsc{answer\_basin}
      & |b| \ge 6,\; \operatorname{medpos}(b) > q_{0.5}^{\text{basin}},\;
        \pi_b \ge 0.6,\; n_{\mathrm{runs}}(b) \ge 3 \\
    \textsc{weak\_basin}
      & |b| \ge 6,\; \operatorname{medpos}(b) > q_{0.5}^{\text{basin}}
        \text{ but answer-basin criteria fail} \\
    \textsc{decision\_point}
      & |b| \le 5 \text{ and touches an articulation point} \\
    \textsc{intermediate}
      & \text{otherwise.}
  \end{cases}
  \label{eq:role}
\end{equation}

\paragraph{Bridge-vs-cycle convention.}
The biconnected decomposition partitions the graph into trivial
$K_2$ bridge blocks and non-trivial blocks of size $|b|\ge 3$ that
are $2$-vertex-connected.  All main-text constructs operate on the
non-trivial subset because including bridges inflates
length-driven fragmentation while adding little semantic content to
the atlas.

\paragraph{Primary-block tie breaks.}
When a slice is covered by multiple non-trivial blocks, the primary
block $b_{r,t}^{\star}$ is chosen by largest block size in nodes, with
ties broken by larger unique-run count, earlier minimum slice
position, then smaller block id.  Slices with no non-trivial block are
skipped in typed-state trajectories rather than mapped to a sentinel.

\subsection{Reproducibility and Statistics}
\label{app:details-trailer}
\label{app:stats}

Fresh CoT generation uses framework-level nondeterminism rather than
fixed per-run sampling seeds, so bitwise reruns of the analysis are
supported by the generated cache and graph summaries; the paper and code repository provide full algorithmic details and exported headline tables.  The paper reports problem-clustered bootstrap
intervals ($B{=}1000$) for headline proportions and means where
appropriate, and null-controlled $z$-scores for graph or
transition-structure diagnostics.  Multiple related null-based tests
within the same experiment family use Benjamini--Hochberg FDR control.

\paragraph{Model checkpoints.}
All measured systems are open-weight Hugging Face checkpoints
(Table~\ref{tab:model-checkpoints}); no fine-tuning is applied and all
generations are zero-shot CoT with each model's default chat template.

\begin{table}[h]
\caption{Model checkpoints used for CoT generation.}
\label{tab:model-checkpoints}
\centering\scriptsize
\begin{tabular}{@{}p{0.14\linewidth}p{0.46\linewidth}p{0.20\linewidth}p{0.10\linewidth}@{}}
\toprule
Paper name & HuggingFace ID & Mode / setting & Arch. \\
\midrule
R1-8B
  & \texttt{deepseek-ai/DeepSeek-R1-0528-Qwen3-8B}
  & thinking-only
  & Qwen3 \\
Think
  & \texttt{Qwen/Qwen3-4B-Thinking-2507}
  & thinking-only
  & Qwen3 \\
Inst
  & \texttt{Qwen/Qwen3-4B-Instruct-2507}
  & non-thinking-only
  & Qwen3 \\
Llama-8B
  & \texttt{deepseek-ai/DeepSeek-R1-Distill-Llama-8B}
  & thinking-only
  & Llama-3.1 \\
Qwen3-32B
  & \texttt{Qwen/Qwen3-32B}
  & \texttt{enable\_thinking=True}
  & Qwen3 \\
Q2.5-72B
  & \texttt{Qwen/Qwen2.5-Math-72B-Instruct}
  & default
  & Qwen2.5 \\
\bottomrule
\end{tabular}
\end{table}

\begin{table}[h]
\caption{CoT generation protocol.  We intentionally use a uniform
sampling protocol across all measured models for atlas comparability,
rather than each model card's recommended decoding parameters;
model-specific chat templates and mode switches are still respected.}
\label{tab:gen-protocol}
\centering\footnotesize
\begin{tabular}{@{}lp{0.68\linewidth}@{}}
\toprule
$N$ per problem-model cell & $64$ \\
Temperature                 & $0.7$ \\
Top-$p$                     & $0.9$; uniform across all models \\
Top-$k$                     & disabled \\
\texttt{max\_tokens}        & $32{,}768$ \\
Chat template               & each model's default chat template applied via \texttt{tokenizer.apply\_chat\_template}; no system prompt \\
User prompt (math)          & \texttt{Return your final response within \textbackslash boxed\{\}.\textbackslash n Question:\textbackslash n\{problem\}} \\
User prompt (GPQA)          & \texttt{Please answer the following multiple choice question:\textbackslash n\textbackslash n Question: \{question\}\textbackslash n\textbackslash n Options:\textbackslash nA.~\{A\}~~B.~\{B\}~~C.~\{C\}~~D.~\{D\}\textbackslash n\textbackslash n Return your final response within \textbackslash boxed\{\} and only include the letter choice (A, B, C, or D) as your final response.} \\
Answer extraction           & parse last \texttt{\textbackslash boxed\{...\}} via brace-matched extraction; numeric benchmarks verified with \texttt{math\_verify} (symbolic equivalence); GPQA verified by exact single-letter match \\
Sampling randomness         & framework-level nondeterminism; analysis reruns use distributed caches \\
\midrule
Mean / median output tokens & $10{,}208$ / $7{,}578$ \\
Length-truncated runs        & $4{,}210/61{,}052 = 6.9\%$ (primary corpus; retained, counted as non-correct if no answer extracted) \\
\bottomrule
\end{tabular}
\end{table}

\paragraph{Hardware and compute.}
CoT generation: $8\times$ NVIDIA H20-96GB GPUs, using
flash-attention-2; the 72B model was served with $8$-way tensor
parallelism, smaller models on a single H20.  Estimated total
generation cost: ${\sim}200$ GPU-hours across all models and benchmarks.
Activation-cache extraction ran on the same hardware during generation.
The SliceGraph analysis pipeline (graph construction, BCC decomposition,
reward field, and typed-state diagnostics) runs entirely on CPU and
completes in $2$--$5$\,s per problem at $N{=}64$ runs.

\paragraph{Software versions.}
Python~3.12; PyTorch~2.6; Transformers~5.3; NetworkX~3.6
(BCC decomposition via \texttt{biconnected\_components}, Louvain via
\texttt{louvain\_communities}); NumPy~2.2; SciPy~1.17;
scikit-learn~1.8 (PCA ablation);
\texttt{sentence-transformers}~(MiniLM ablation).

\paragraph{Random seeds.}
Louvain community detection uses seed $42$; problem-clustered bootstrap
uses $B{=}1{,}000$ resamples with \texttt{numpy.random.seed(0)} per
bootstrap call; graph-null rewiring uses $3$ independent seeds per
variant.  Temperature-sampled CoT runs have no fixed seed (each run is
an independent sample from the model's output distribution).

\section{Representation and Family Robustness}
\label{app:robustness}

\subsection{Block-Scale Sensitivity and Hyperparameters}
\label{app:scale-sensitivity}

The paper uses the graph-theoretic split between trivial BCCs
($K_2$, bridges) and non-trivial biconnected blocks ($|b|\ge 3$),
i.e.\ $2$-vertex-connected cyclic regions.  Moving from the full
decomposition ($m{=}2$) to the non-trivial convention ($m{=}3$)
changes the object substantially: the mean retained block count drops
from $575.2$ to $49.5$.  Even so, the family-level atlas picture is
stable.  Including bridges shifts the multi-family headline only
slightly ($85.5\%\to 86.8\%$) but overfragments the partition, raising
mean correct-only families per cell from $4.66$ to $11.27$ and the
between-family length-variance ratio from $0.460$ to $0.697$.  We
therefore retain bridges as connectors but not as semantic regions.

Across a $30$-configuration sweep over
$(\texttt{sep\_up},k,\texttt{mutual})$ the default configuration is
not a narrow optimum, and the multi-family rate of
\S\ref{sec:multi-route-geometry} remains stable across all sampled
settings.  On run-number sufficiency, $N{=}64$ retains $95.0\%$ of the
process-family coverage observed on deeper $N{=}256$ caches.

\paragraph{RBF kernel scale $\sigma$.}
The edge-weight scale $\sigma\in\{0.20, 0.35, 0.50\}$ has no effect
on topology-dependent readouts because the mutual-$k$NN construction
determines graph topology independent of weight scaling: $\sigma$
modulates edge strength but not edge existence.  On the sweep-eligible
subset, the multi-family rate remains $97.8\%$ and the pairwise isomer
rate remains $76.6\%$ at all three values.

\paragraph{Reward-field diffusion $\alpha$ and core quartile $q$.}
Varying the PageRank damping $\alpha\in\{0.50,0.65,0.80\}$ and core
extraction percentile $q\in\{0.50,0.75,0.90\}$ preserves the
correct-run field advantage in all configurations: non-canonical
variants produce equal or larger correct-vs-incorrect separation on
field-score and entry-progress metrics across all three models.  The
canonical $\alpha{=}0.65$, $q{=}0.75$ is therefore a
\emph{conservative} choice; no variant degrades the reward-field
signal or inverts the multi-core structure.

\paragraph{Role thresholds.}
An OAT sweep over $8$ role parameters ($46$ configurations including
default) shows sign stability for the headline family-TV-above-null
direction across all $19$ non-degenerate configurations.
Family-TV does not affect family construction, which uses block
co-visitation.  The typed-state Family-TV diagnostic does use role
labels, but the OAT sweep shows its sign and qualitative conclusions
are stable under all non-degenerate role-threshold variants.

\subsection{Representation Ablation: Per-(Pipeline$\times$Model)}
\label{app:repr-ablation}

The full representation ablation discussed in
\S\ref{sec:representation-validity} is reported here.  Pipelines are
mutual-$k$NN graphs with the same BCC decomposition and vary only the
slice representation: cosine on the same activation keys; PCA-cosine
at $d\in\{2,10,50\}$; and \texttt{all-MiniLM-L6-v2} sentence
embeddings on $256$-token slice text.  ``collapsed'' indicates a graph
too degenerate for a useful BCC decomposition.  We interpret the
MiniLM failure as a same-problem saturation effect: within one
problem, slice text embeddings preserve topical relatedness but
compress many slices into a narrow similarity range, leaving too
little within-problem variance for mutual-$k$NN to separate local
trunks, forks, and re-joins.

\begin{table}[h]
\caption{Representation ablation, full per-(pipeline$\times$model)
family geometry breakdown: multi-fam~\% / mean correct families per
cell on the same graphs as Table~\ref{tab:repr-ablation-main} of the
main text.}
\label{tab:repr-ablation}
\centering\footnotesize
\begin{tabular}{@{}lccc@{}}
\toprule
Pipeline & R1-8B & Think & Inst \\
\midrule
\textbf{Jaccard kNN (ours)}
  & $\mathbf{89.3 / 5.14}$
  & $\mathbf{86.2 / 4.61}$
  & $\mathbf{81.1 / 4.21}$ \\
Cosine kNN
  & $88.1 / 4.80$
  & $85.5 / 4.36$
  & $79.6 / 3.87$ \\
Overlap kNN
  & $87.7 / 4.77$
  & $85.5 / 4.36$
  & $79.6 / 3.87$ \\
PCA-$50$ + cos
  & $81.4 / 3.38$
  & $81.4 / 3.33$
  & $73.8 / 3.00$ \\
PCA-$10$ + cos
  & $84.9 / 3.45$
  & $81.6 / 3.40$
  & $73.5 / 2.96$ \\
PCA-$2$ + cos
  & $83.3 / 3.65$
  & $82.1 / 3.76$
  & $74.1 / 3.52$ \\
MiniLM slice
  & \emph{collapsed} ($n{=}5$)
  & \emph{collapsed} ($n{=}5$)
  & \emph{collapsed} ($n{=}11$) \\
\bottomrule
\end{tabular}
\end{table}

\subsection{Fixed-Scaffold Split-Half Stability}
\label{app:split-half}

As a lightweight held-out-style check, we split each cell's runs into
two deterministic $32/32$ halves after constructing the canonical
SliceGraph and non-trivial block scaffold.  Within each half we induce
the run-level co-visitation graph on normalized-correct runs only and
rerun the same Louvain family detector.  This does not yet prove that a
graph built from one half can project every held-out run into a matched
family, but it tests whether the observed family multiplicity depends on
a small set of runs that vanish under resampling.

Across $779$ non-code primary cells with at least two correct runs in
both halves, both halves are multi-family in $97.2\%$ of cells
(problem-clustered bootstrap $95\%$ CI $[96.0,98.2]\%$).  The full-cell
multi-family rate on this split-valid subset is $99.7\%$.  The mean
absolute difference in half-level family count is $1.12$ families
(median $1$), and the two half-level pairwise family-isomer rates are
$75.1\%$ and $75.7\%$, with median absolute pair-rate difference
$5.1$ points.

We treat this as a lightweight stability check.  The stricter held-out
validation below completes the picture.

\subsection{Held-Out Atlas Validation}
\label{app:heldout-atlas}

To move beyond a fixed-scaffold stability check, we perform a fully
held-out validation: (i) build a SliceGraph from a random $32$-run
\emph{train half}; (ii) for each held-out slice, compute nearest neighbours against
train-half slices only using its sparse activation keys (without
test-test edges or the full $64$-run graph topology) and assign it to
that neighbour's primary non-trivial block; (iii) compute covisitation
families on the held-out correct runs using the train-half block
scaffold; and (iv) measure whether the held-out partition is
multi-family and how it relates to the full-sample partition.

We run $R{=}5$ random splits per cell across $834$ canonical primary
cells with ${\ge}4$ correct runs.  Block coverage from test onto
train blocks is $\mathbf{95.5\%}$, confirming that the scaffold
generalizes.  Held-out cells are multi-family in
$\mathbf{91.0\%}$ of cases (vs.\ $97.8\%$ on the full $64$-run
sample), with mean held-out pairwise isomer rate $\mathbf{60.1\%}$
(vs.\ $77.8\%$ full-sample).  NMI between full-sample and held-out
family labels is moderate ($0.24$ median), indicating that exact family
identities are not stable under a half-sample scaffold.  We therefore
use this held-out check to support multi-family presence and block
coverage, not one-to-one family recovery.

\subsection{Correct-Run Controlled Isomer Rate}
\label{app:controlled-isomer-rate}

Family count might be confounded by the number of correct runs: cells
with higher accuracy have more correct runs and might discover more
families simply due to sample size.  We control for this by fixing the
number of correct runs to $m\in\{4,8,16\}$, subsampling $m$ correct
runs per cell (with $R{=}10$ replicates) and recomputing the pairwise
isomer rate on the same block scaffold.

\begin{table}[h]
\caption{Pairwise isomer rate at fixed correct-run count $m$.
Mean over $R{=}10$ replicates per cell, pooled across three primary
models.  Multi-family structure persists even at $m{=}4$.}
\label{tab:controlled-isomer}
\centering\footnotesize
\begin{tabular}{@{}lrrr@{}}
\toprule
& $m{=}4$ & $m{=}8$ & $m{=}16$ \\
\midrule
Mean isomer rate & $58.2\%$ & $69.6\%$ & $75.2\%$ \\
Multi-family \%  & $82.1\%$ & $99.7\%$ & $100\%$  \\
Mean families    & $2.3$    & $3.3$    & $4.4$    \\
\bottomrule
\end{tabular}
\end{table}

Even at $m{=}4$, the multi-family rate is $82.1\%$ and the pairwise
isomer rate is $58.2\%$, still substantial.  At $m{=}8$ the
rate already recovers to ${\sim}70\%$, and at $m{=}16$ it nearly
matches the full-sample $76.6\%$ headline.  The claim is therefore not
an artefact of high correct-run counts inflating opportunity.

\subsection{Family Construction Sensitivity}
\label{app:family-construction-sensitivity}

We sweep the two main Louvain hyperparameters---resolution
$\rho\in\{0.5,1.0,1.5,2.0\}$ and weighted-Jaccard edge threshold
$\tau\in\{0.025,0.05,0.1\}$---across $834$ primary cells.

\begin{table}[h]
\caption{Pairwise isomer rate (full-cell block weights) under a
resolution$\times$threshold sweep on the correct-only covisitation
graph.  Canonical configuration: $\rho{=}1.0$, $\tau{=}0.05$.}
\label{tab:family-sensitivity}
\centering\footnotesize
\begin{tabular}{@{}lrrr@{}}
\toprule
Resolution & $\tau{=}0.025$ & $\tau{=}0.05$ & $\tau{=}0.1$ \\
\midrule
$0.5$ & $0.247$ & $0.365$ & $0.552$ \\
$\mathbf{1.0}$ & $0.744$ & $\mathbf{0.766}$ & $0.793$ \\
$1.5$ & $0.857$ & $0.865$ & $0.868$ \\
$2.0$ & $0.917$ & $0.915$ & $0.913$ \\
\bottomrule
\end{tabular}
\end{table}

At $\rho{=}1.0$ the isomer rate is stable across thresholds ($74$--$79\%$),
and the multi-family rate exceeds $96\%$ for all three threshold values.
Lower resolution ($\rho{=}0.5$) under-partitions and higher resolution
($\rho{\ge}1.5$) over-partitions, but the multi-family presence is
robust across $\rho{\ge}1.0$.  Replacing full-cell block coverage
weights with correct-only weights changes the canonical isomer rate by
only $3.4$ percentage points ($76.6\%\to73.2\%$), confirming that
incorrect runs' contribution to rare-block upweighting is not the driver.

\subsection{Isomer-Discovery Curves under Run Subsampling}
\label{app:isomer-discovery-curves}

We subsample the existing $N{=}64$ caches at $N\in\{8,16,32,64\}$
($R{=}5$ replicates, uniform without replacement) and recompute family
detection and core masks at canonical hyperparameters.

\begin{figure}[h]
\centering
\includegraphics[width=\linewidth]{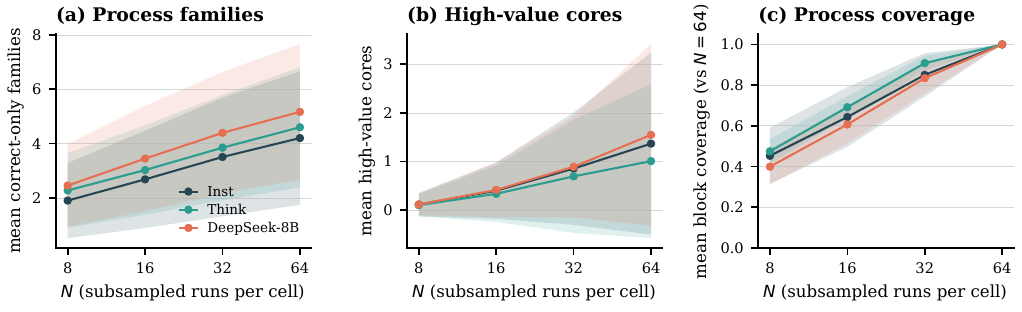}
\caption{Isomer-discovery curves under run subsampling.
(a)~Correct-only family count, (b)~high-value-core components,
(c)~block coverage relative to $N{=}64$.
Block coverage saturates around $N{\approx}32$; family and core
counts keep growing as rare-block neighbourhoods enter the sample.}
\label{fig:isomer-discovery-curves}
\end{figure}

\section{Process-Isomer Geometry and Dynamics}
\label{app:reward-detail}
\label{app:reward-and-dynamics}

\begin{table*}[t]
\caption{Full model$\times$dataset route multiplicity including replication models.
Entries: \% cells with ${\ge}2$ correct-only families\,/\,mean correct families.
Top: primary corpus; bottom: replication models.
Llama-8B covers AIME24, AIME25, and GPQA only.}
\label{tab:model-dataset-full}
\centering\footnotesize
\begin{tabular}{@{}lcccccc@{}}
\toprule
Model & AIME24 & AIME25 & BRUMO25 & GPQA & HMMT25 & All \\
\midrule
R1-8B  & 86.7\,/\,4.87 & 86.7\,/\,4.80 & 86.7\,/\,4.57 & 92.4\,/\,5.50 & 76.7\,/\,3.97 & 89.3\,/\,5.14 \\
Think        & 93.3\,/\,5.07 & 86.7\,/\,5.10 & 96.7\,/\,4.77 & 85.9\,/\,4.60 & 70.0\,/\,3.57 & 86.2\,/\,4.61 \\
Inst         & 83.3\,/\,4.20 & 70.0\,/\,3.43 & 76.7\,/\,3.70 & 87.9\,/\,4.72 & 50.0\,/\,2.17 & 81.1\,/\,4.21 \\
\midrule
Llama-8B     & 76.7\,/\,4.80 & 53.3\,/\,3.10 &               & 82.3\,/\,3.98 &               & 78.3\,/\,3.98 \\
Qwen3-32B    & 93.3\,/\,5.17 & 90.0\,/\,4.93 & 96.7\,/\,4.80 & 88.4\,/\,4.56 & 86.7\,/\,4.17 & 89.6\,/\,4.64 \\
Q2.5-72B     & 46.7\,/\,1.93 & 36.7\,/\,1.33 & 46.7\,/\,2.40 & 86.4\,/\,4.18 & 13.3\,/\,0.47 & 67.3\,/\,3.18 \\
\bottomrule
\end{tabular}
\end{table*}

The reward field is used in this paper only as a
\emph{label-seeded explanatory solution landscape}.  A cell enters the
reward-evaluable subset only when the leave-one-out seed is stable
enough to define such a landscape; this is a construction requirement,
not a classification objective.

\paragraph{Reward-field propagation details.}
For a cell with $n$ graph-valid runs and block visitor set
$\mathcal R_b$, the global seed used for high-value cores is
\begin{equation}
  v_b^{(0)} =
  \mathbf 1[3\le n_b\le n-3]
  \left(\frac{1}{n_b}\sum_{r\in\mathcal R_b} y_r -
  \frac{1}{n}\sum_r y_r\right)
  \left(\frac{n_b}{n}\right)^{1/2},
  \qquad n_b=|\mathcal R_b|.
  \label{eq:reward-seed-app}
\end{equation}
The run-level leave-one-out variant used for per-run overlays removes
that run from both the block and cell denominators.  The diffusion graph
has non-trivial blocks as nodes; $K_2$ bridges are not diffusion nodes.
Let $E_{\mathrm{BC}}$ connect two non-trivial blocks that share an
articulation point in the block-cut tree, and let $E_{\mathrm{temp}}$
connect consecutive distinct primary blocks in a compacted run path.
Temporal edges are symmetrized, multiplicities are binarized, and the
raw adjacency is
\begin{equation}
  A^{\mathrm{raw}}_{ij}=\mathbf 1[i=j] +
  \mathbf 1[(i,j)\in E_{\mathrm{BC}}\cup E_{\mathrm{temp}}],
  \qquad
  P_{ij}=\frac{A^{\mathrm{raw}}_{ij}}{\sum_k A^{\mathrm{raw}}_{ik}} .
  \label{eq:reward-adjacency-app}
\end{equation}
Thus block-cut and temporal edges have equal weight ($\lambda=1$) after
binarization, and the self-loop weight is $1$.  In all experiments we
use $\alpha{=}0.65$ for $24$ steps,
\begin{equation}
  \mathbf{v}^{(t+1)} =
  \alpha\,\mathbf{v}^{(0)} + (1{-}\alpha)\,\mathbf{P}\mathbf{v}^{(t)}.
  \label{eq:reward-field}
\end{equation}
We keep $\mathbf v^{(24)}$ as the field used throughout
\S\ref{sec:value-landscape}; the high-value core is the top
quartile of its strictly positive support.
After $24$ iterations, each propagated field---pooled or
family-conditioned---is independently rescaled by its own $\ell_\infty$
norm.  Core masks, connected components, and specialization statistics are
invariant to positive rescaling; field-sharpness ratios
(Equation~\ref{eq:field-sharpness-app}) should be read under this per-field
normalization convention.

\paragraph{Alignment and non-redundancy definitions.}
For a cell with high-value-core components $C_1,\dots,C_K$ and a
correct-only family $f$, let $B_f$ be the union of non-trivial blocks
visited by runs in $f$ and
$m_f(k)=|B_f\cap C_k|/|C_k|$.  For $K\ge2$ we normalize each non-zero
row to $p_f(k)=m_f(k)/\sum_j m_f(j)$ and define
\begin{equation}
  \mathrm{Spec}=1-\frac{1}{|F_+|}\sum_{f\in F_+}
  \frac{H(p_f)}{\log K},
  \label{eq:family-core-spec-app}
\end{equation}
where $F_+$ are families with non-zero core footprint.  The reported
$0.59$ is averaged on the $324$ multi-core reward-evaluable cells.
For a family-conditioned field $\mathbf v_f$, computed by setting
correctness labels outside family $f$ to zero before applying
Equations~\ref{eq:reward-seed-app}--\ref{eq:reward-field}, field
sharpness is
\begin{equation}
  \mathrm{Sharp}(f)=\frac{\max_i v_{f,i}^{(24)}}{\max_i v_i^{(24)}} .
  \label{eq:field-sharpness-app}
\end{equation}
Coverage loss uses the most coverage-critical family in a cell:
\begin{equation}
  \Delta_{\max}=\max_f
  \frac{|(\cup_g B_g)\setminus(\cup_{g\ne f}B_g)|}{|\cup_g B_g|}.
  \label{eq:coverage-loss-app}
\end{equation}
The $94.5\%$ headline is $\Pr(\Delta_{\max}>0.10)$ over the $816$
canonical cells with at least two correct-only families.

\begin{table}[h]
\caption{Multi-modal reward geometry (non-code).  Sections A--C use
the three primary models; section D adds the Llama cross-architecture
replication.}
\label{tab:multi-modal}
\centering\footnotesize
\begin{tabular}{@{}llc@{}}
\toprule
  & Statistic & Value \\
\midrule
\multicolumn{3}{@{}l}{\emph{A. High-value core disconnectedness ($n{=}479$)}} \\
  & \% cells with $\ge 2$ core components      & $67.6\%$ \\
  & Mean / median $\#$ core components            & $2.61 / 2$ \\
  & Size ratio: 2nd / 1st component               & $0.54$  \\
\midrule
\multicolumn{3}{@{}l}{\emph{B. Correct-only families ($n{=}954$)}} \\
  & \% cells with $>1$ correct family          & $85.5\%$ \\
  & Mean correct families / cell               & $4.66$   \\
  & Pairwise family-isomer rate ($n{=}834$ cells) & mean $76.6\%$, median $79.6\%$ \\
  & Problem-clustered bootstrap CI for mean       & $[75.6,77.5]\%$ \\
  & Split-half both-halves multi-family ($n{=}779$ cells) & $97.2\%$ \\
  & Length-variance ratio                         & $0.460$  \\
  & Modularity null $z$ (median)                  & $35.54$  \\
\midrule
\multicolumn{3}{@{}l}{\emph{C. Alignment and non-redundancy (natural denominators)}} \\
  & Family--core footprint specialization (multi-core cells, $n{=}324$) & $0.59$   \\
  & Family field-max / pooled-global field-max ($2{,}094$ family fields) & $1.55\times$ \\
  & Jaccard(family core, global core)             & $0.17$   \\
  & Spearman($\#$components, $\#$families)        & $0.08$   \\
  & \% multi-family cells where most critical family removal loses $>10\%$ & $94.5\%$ \\
  & Core-divergence increment over family-isomer rate & $+1.3$\,pp \\
\midrule
\multicolumn{3}{@{}l}{\emph{D. Llama cross-architecture replication}} \\
  & \% with $\ge 2$ core components ($n{=}194$)  & $78.4\%$ \\
  & \% with $>1$ correct family ($n{=}258$)       & $78.3\%$ \\
  & Family--core footprint specialization (multi-core) & $0.60$ \\
  & Family field-max / pooled-global field-max    & $1.46\times$ \\
  & Length-variance ratio                         & $0.469$ \\
\bottomrule
\end{tabular}
\end{table}

The pairwise family-isomer rate is the primary Definition~\ref{def:isomer}
rate: it is computed within each problem-model cell that has at least two
normalized-correct runs, as the share of unordered correct-run pairs
assigned to different correct-only process families, then averaged across
cells.  Core divergence is reported separately as a value-landscape
refinement (\S\ref{sec:value-landscape}).

\subsection{Per-Model Block-Type Localisation}
\label{app:reward-field-per-model}

The pooled reward-field statistics collapse a model-dependent
block-type localisation that is itself informative.  Inst
concentrates at \textsc{shared\_trunk}; R1-8B, Think, and
Llama-8B concentrate at \textsc{answer\_basin}.  All four therefore
exhibit a localized high-value landscape; they differ only in which
structural niche hosts it.

\begin{table}[h]
\caption{Per-model mean propagated reward field by block-type role on
the canonical reward-field corpus.  Bold cells indicate the dominant
block-type per model.}
\label{tab:reward-field-block-type}
\centering\footnotesize
\begin{tabular}{@{}lrrrrr@{}}
\toprule
Model & shared\_trunk & answer\_basin & weak\_basin & intermediate & decision\_point \\
\midrule
Inst        & $\mathbf{0.138}$ & $0.001$ & $0.016$ & $0.005$ & $0.003$ \\
R1-8B & $0.062$          & $\mathbf{0.252}$ & $0.062$ & $0.029$ & $0.023$ \\
Think       & $0.064$          & $\mathbf{0.236}$ & $0.046$ & $0.030$ & $0.023$ \\
Llama-8B    & $0.031$          & $\mathbf{0.123}$ & $0.044$ & $0.006$ & $0.006$ \\
\bottomrule
\end{tabular}
\end{table}

\subsection{Family-Specific Navigation Diagnostics}
\label{app:reward-dynamics-table}

The typed-state transition model is used to compare how different
correct-only families traverse the same atlas.  The population-level
dynamics headline in the atlas version is family-specific kernel
divergence rather than an auxiliary task objective.

For each run we map the primary non-trivial block at slice $t$ to
$z_{r,t}=(\operatorname{role}(b_{r,t}^{\star}),\operatorname{posbin}(\tau_t),
\operatorname{core}(b_{r,t}^{\star}))$, compact consecutive duplicates,
and append absorbing correct/wrong terminals.  Transition counts $C$
are Laplace-smoothed into
\begin{equation}
  P_{ij}=\frac{C_{ij}+\alpha_{\mathrm{smooth}}}
  {\sum_{j'} C_{ij'}+\alpha_{\mathrm{smooth}}|\mathcal S|},
  \qquad \alpha_{\mathrm{smooth}}=0.5.
  \label{eq:typed-kernel-app}
\end{equation}
Family-TV averages row-wise kernel divergence,
\begin{equation}
  \mathrm{TV}(P_f,P_{f'}) =
  \frac{1}{2|\mathcal S|}\sum_{i\in\mathcal S}
  \|P_f(i,\cdot)-P_{f'}(i,\cdot)\|_1.
  \label{eq:family-tv-app}
\end{equation}
In the released implementation, the averaging set $\mathcal S$ includes the
two absorbing terminal rows ($\mathrm{EOS}_{\mathrm{correct}}$,
$\mathrm{EOS}_{\mathrm{wrong}}$); because both are forced absorbing with
identical rows across families, they contribute zero TV\@.  Removing them
rescales the denominator but does not affect the matched-null comparison
reported in \S\ref{sec:dynamics}.
The pooled-kernel committor solves $(I-Q)q=r_A$, where $Q$ is the
transient-to-transient block and $r_A$ is the transition vector into
$\mathrm{EOS}_{\mathrm{correct}}$.  The three-step escape hazard is
$h_3=\sum_{i\in\mathcal C}\hat\pi(i)
\sum_{j\notin\mathcal C}P^3_{ij}$; return and MFPT readouts use the
same smoothed kernel and are retained as case-level diagnostics.

\begin{table}[h]
\caption{Family-specific navigation diagnostics on the
core-eligible multi-family subset ($n{=}403$).}
\label{tab:family-tv}
\centering\footnotesize
\begin{tabular}{@{}lc@{}}
\toprule
Statistic & Value \\
\midrule
Median family pairwise kernel TV & $0.137$ \\
\% cells with real TV $>$ null $p_{95}$ & $80.9\%$ \\
Median null $z$ & $3.14$ \\
Corr.(family-TV, static specialization) & $-0.07$ \\
\bottomrule
\end{tabular}
\end{table}

\begin{table}[h]
\caption{Common-support family-TV diagnostic.  TV is averaged only over
typed-state rows visited by both families in a pair; the null shuffles
run-to-family assignments while preserving family sizes.}
\label{tab:family-tv-common-support}
\centering\footnotesize
\begin{tabular}{@{}lc@{}}
\toprule
Statistic & Value \\
\midrule
Eligible cells & $403$ \\
Median common-support family TV & $0.206$ \\
Median common-support null $p_{95}$ & $0.183$ \\
\% cells with common-support TV $>$ null $p_{95}$ & $77.4\%$ \\
Median common-support null $z$ & $3.45$ \\
Median shared typed states per family pair & $7.0$ \\
\bottomrule
\end{tabular}
\end{table}

The pooled-kernel committor, MFPT to core, escape hazard, and
post-escape return remain useful \emph{case-level atlas readouts} for
understanding how one family navigates relative to another.  In this
atlas version they are not treated as standalone population claims.

\subsection{Typed-State Sensitivity and Scope}
\label{app:dynamics-alpha-note}

Family-TV is the only population-level dynamics statistic we headline,
because it is the most stable to mild changes in smoothing and role
thresholds.  The companion pooled committor, MFPT, escape, and return
diagnostics remain informative for inspecting one fixed atlas, but
their absolute numeric values are more sensitive to role-threshold
choices and sparse-transition smoothing.  We therefore use them as
descriptive overlays on the process atlas rather than as independent
paper claims.

\subsection{Null-Control Ledger}
\label{app:dynamics-graph-null}

Table~\ref{tab:claim-ledger} maps each main-text claim to the
corresponding weak reading and diagnostic used to control it.

\begin{table}[h]
\centering\small
\caption{Claim-to-control ledger for the main atlas readouts.}
\label{tab:claim-ledger}
\begin{tabular}{@{}lll@{}}
\toprule
Claim & Weak reading & Control / diagnostic \\
\midrule
Families are route units & arbitrary graph clusters & human strategy audit, modularity null \\
Process isomers are common & few-run sampling artifact & pairwise rate, split-half stability \\
High-value cores are multi-basin & heatmap fragmentation & core-size ratio, label-permutation caveat \\
Family dynamics differ & support mismatch only & family-label null, common-support TV \\
\bottomrule
\end{tabular}
\end{table}

The paper uses several nulls, each aimed at a different possible
artifact.  We use five null families, each targeting a different
possible confound:
\begin{enumerate}[nosep,leftmargin=*]
\item \textbf{Degree-preserving graph rewire} --- preserves the degree
      sequence but destroys local adjacency; used for the family
      modularity headline (median $z{=}35.54$).
\item \textbf{Block-type-preserving rewire} --- preserves role counts
      and type marginals; used as a structural-sensitivity stress test
      (effect-drop only).
\item \textbf{Family-label shuffle} --- preserves family sizes and
      typed-state support but destroys route-specific labels; the
      exported population null for family-TV ($80.9\%$ above $p_{95}$,
      median $z{=}3.14$).
\item \textbf{Temporal-order shuffle} --- preserves visited typed
      states but destroys transition order; used for kernel, escape,
      return, and MFPT stress tests (effect-drop only).
\item \textbf{Label permutation} --- preserves graph topology, family
      partition, and per-cell label count but randomises correctness
      association; the reward-core stress test
      (Table~\ref{tab:label-permutation-reward-null}).
\end{enumerate}
\label{tab:null-ledger}%
Items~1 and~3 generate the exported headline nulls; items~2 and~4 are
validity-ladder stress tests and should not be read as independent
population claims.

\newcommand{\rewardnullrange}[2]{[#1,#2]}
\begin{table}[h]
\caption{Label-permutation reward-field stress test.  Permutations preserve the graph topology, family partition, and the number of correct labels in each reward-evaluable cell, but randomize which runs are correct.  The permutation range reports shuffle-level $p_5$--$p_{95}$ across $100$ shuffles.}
\label{tab:label-permutation-reward-null}
\centering\footnotesize
\begin{tabular}{@{}p{0.42\textwidth}ccc@{}}
\toprule
Readout & Real & Label-perm. mean \rewardnullrange{$p_5$}{$p_{95}$} & $z$ \\
\midrule
Multi-core rate & $67.6\%$ & $85.4\%$ \rewardnullrange{$83.3$}{$88.1$}$\%$ & $-11.44$ \\
Family--core footprint specialization, real multi-core cells & $0.59$ & $0.74$ \rewardnullrange{$0.72$}{$0.76$} & $-11.87$ \\
Family field-max / pooled-global field-max, field-weighted & $1.55\times$ & $1.55\times$ \rewardnullrange{$1.48$}{$1.70$} & $0.08$ \\
Jaccard(family core, global core), field-weighted & $0.17$ & $0.10$ \rewardnullrange{$0.097$}{$0.105$} & $29.36$ \\
\bottomrule
\end{tabular}
\end{table}

Label permutations tend to scatter positive seed mass, which can create
more disconnected random cores and higher apparent specialization than
the real labels.  We therefore do not use this stress test as a
one-sided null for multi-core counts.  Its main diagnostic value is that
real family-conditioned cores overlap the real global core far more than
permuted labels do, while field sharpness alone is comparable to the
permutation range and is interpreted only together with the family/core
geometry above.

\subsection{Family-TV Null Construction}
\label{app:family-tv-null}

For each core-eligible multi-family cell, the family-TV null keeps
family sizes and typed-state support fixed while shuffling family
labels across trajectories before re-estimating per-family kernels.
This preserves the atlas and its coarse traffic levels while erasing
route-specific navigation structure.  The main-text
$80.9\%$-above-null and median-$z{=}3.14$ headline is measured against
this null.

\paragraph{Common-support TV.}
A stricter diagnostic averages TV only over typed states visited by both
families in a pair.  This tests whether families move differently even
where they share support, rather than only because they visit different
states.  Table~\ref{tab:family-tv-common-support} reports the matched
family-size shuffle null and the shared-support denominator summary.

\subsection{Worked Route Report on GPQA-71}
\label{app:route-report}

As a worked example we pick GPQA-71 under \textbf{Inst}
(Qwen3-4B-Instruct-2507): seven correct-only process families
covering $38$ non-trivial blocks across four disconnected high-value-core
components.  Removing the single most coverage-critical family loses
$55.3\%$ of total block coverage, and the loss-if-removed shares sum
to $92.1\%$, confirming that route coverage is family-unique rather
than redundantly shared.  The four core components are occupied by
different families (F6$\to$core~0, F0$\to$core~1, F2/F5$\to$core~2/3),
while F1, F3, F4 are off-core under the top-quartile mask.
Per-family coverage plots and the full summary card are included in
the code repository.

\section{Annotation Protocols and Family Robustness}
\label{app:family-robustness}
\label{app:annotation-cross}

This section covers family-detection robustness and the two
human-annotation tasks used in \S\ref{sec:semantic-validity}.
Task~A audits BCC shared-state semantics; Task~B audits process-family
same-strategy semantics.

\begin{table}[h]
\caption{Compact summary of annotation and family-validation
statistics, split by task.  Top block: Task~B (process families);
bottom block: Task~A (BCC shared-state).}
\label{tab:appendix-families}
\centering\footnotesize
\begin{tabular}{@{}lc@{}}
\toprule
Statistic & Value \\
\midrule
\multicolumn{2}{@{}l}{\emph{Task~B (process families)}} \\
Within-family decisive agreement (Annotator~A) & 94.4\% \\
Self-consistency $\kappa$ (Annotator~A) & 0.65 \\
Across-family cross-annotator $\kappa$ (A $\times$ B) & \textbf{0.737} \\
Decisive exact agreement, family pack (A $\times$ B, $n{=}55$) & 81.8\% \\
\midrule
\multicolumn{2}{@{}l}{\emph{Task~A (BCC shared-state)}} \\
BCC agreement vs.\ ground-truth (Annotator~B, $n{=}72$) & \textbf{90.3\%} \\
BCC agreement vs.\ ground-truth (Annotator~A, $n{=}72$) & 88.9\% \\
BCC-pack cross-annotator $\kappa$ (A $\times$ B, decisive $n{=}71$) & \textbf{0.97} \\
Matched non-articulation specificity (Annotator~B) & $36/36$ \\
\bottomrule
\end{tabular}
\end{table}

\subsection{Family Detection Robustness}
\label{app:family-detection-robust}

The default $85.5\%$ multi-family rate is not Louvain-specific:
greedy modularity reproduces the same direction and magnitude, while a
degree-preserving modularity null is decisively weaker (median
$z{=}35.54$).  Label propagation is included only for transparency and
is not preferred because it collapses weighted-Jaccard family graphs
pathologically.

\subsection{Annotation Protocol}
\label{app:annotation-protocol}

Annotators~A and~B are two non-author external annotators with
mathematical reasoning and LLM-trace experience.  They are blind both
to the project hypothesis and to within/across pair labels, and
cross-label both the BCC pack (Task~A) and the process-family pack
(Task~B) under matched strict rubrics.

\paragraph{Pair construction.}
Task~A (BCC): for each BCC block, a within-BCC pair is formed by
selecting two runs from different trajectories with the smallest
positional gap; a matched across-BCC control with the same correctness
label and the smallest absolute position offset is paired.  For
articulation transitions, the first positive articulation-boundary
crossing per trajectory is matched with a phase-matched
non-articulation control.
Task~B (families): $8$ cases are selected across $3$ models and $3$
datasets, each with ${\ge}4$ correct-only families and
${\ge}5$ decision forks; $2$ runs per family are sampled from
normalized-correct runs only, balanced by length and model/dataset via
stratified sampling, yielding $46$ within-family + $31$ across-family
pairs from $102$ runs.

\paragraph{Blinding and randomisation.}
The self-consistency re-review uses a strictly blinded prompt template
that hides: (i) within-vs-across family labels, (ii) original
first-pass judgements, (iii) family IDs.  Pair order is randomised.
Pair identity is tracked by a deterministic $12$-character hash of
(case\_id, sorted run IDs).  The cross-annotator wave uses the same
blinded pack in the same randomised order, in an independent session
with no access to the primary annotator's responses.

\paragraph{Response schemas.}
Task~A: \texttt{annot\_same\_reasoning\_state} $\in$
\{yes, no, uncertain\}, \texttt{annot\_confidence} $\in$ 1--5, plus
free-text \texttt{annot\_notes}.
Task~B: \texttt{same\_strategy} $\in$ \{yes, no, uncertain\},
\texttt{confidence} $\in$ 1--5, \texttt{justification} (one sentence).

\paragraph{Uncertainty handling.}
All tasks use a three-value scheme (yes/no/uncertain).  The instruction
states: ``Use \texttt{uncertain} sparingly---only when two runs share a
framework but execute it in materially different ways.''  We report two
denominators: \emph{raw} (uncertain counts as negative) and
\emph{decisive} (uncertain excluded from both numerator and
denominator).  The decisive rate is the primary claim metric; raw counts or uncertain counts are reported where they
affect interpretation.

\paragraph{Cohen's $\kappa$ calculation.}
We use unweighted Cohen's $\kappa$ with the label set \{yes, no,
uncertain\}.  For the across-family sub-task, labels are inverted
(\texttt{different\_strategy=yes} $\to$ \texttt{same\_strategy=no})
before computing $\kappa$.  Key values: Annotator~A self-consistency
$\kappa{=}0.65$ (overall, $76$ decisive), across-family-only
$\kappa{=}0.80$; cross-annotator across-family $\kappa{=}0.737$
($30$ decisive pairs).

\subsection{Task A: BCC Shared-State Audit}
\label{app:annotation-bcc}

\paragraph{Protocol.}
The core BCC pack contains $72$ slice-text pairs: $36$ within-BCC
pairs and $36$ length-matched across-BCC controls.  Both annotators
judge whether the two slices represent the same reasoning state.
Separately, we construct an articulation transition pack of $72$
adjacent-slice comparisons ($36$ true articulation crossings and $36$
matched non-articulation controls); this auxiliary pack is used as a
precision-side control rather than as a second headline task.

\paragraph{Cross-annotator BCC agreement.}
On the $72$-pair BCC slice-text pack, Annotator~B agrees with the
BCC-derived labels on $\mathbf{90.3\%}$ of pairs and Annotator~A
reaches $88.9\%$.  Direct A-vs.-B agreement is $\mathbf{97.2\%}$ on
all decisive comparisons ($\kappa{=}\mathbf{0.97}$, $n{=}71$), with
only two decisive disagreements.  This is the strongest evidence that
the BCC/shared-state interpretation is independently recoverable.

\subsection{Task B: Process-Family Same-Strategy Audit}
\label{app:annotation-family-task}

\paragraph{Protocol.}
We select $8$ cases across the $3$ primary models and three datasets
(AIME24, HMMT25, and GPQA), requiring at least $4$
correct-only families, at least $5$ decision forks, and problem
accuracy between $40$--$80\%$.  From these cases we sample $102$ runs,
yielding a final judged pack of $77$ blinded run pairs:
$46$ within-family and $31$ across-family.  Input CoTs are truncated
to $15{,}000$ characters via a front/middle/tail
($5$k/$5$k/$5$k) scheme with omissions marked explicitly.

\paragraph{Cross-annotator family agreement.}
The numbers referenced in the main text are: (i) Annotator~A's
within-family decisive agreement rate of $94.4\%$ ($34/36$ decisive
pairs judged same-strategy); (ii) cross-annotator $\kappa{=}0.737$ on
the across-family slice ($n{=}30$ decisive pairs, raw agreement
$90.0\%$); and (iii) $81.8\%$ exact agreement on the subset where
both annotators give a decisive judgement ($n{=}55$).

\paragraph{Across-family differentiation rate.}
Of $31$ across-family pairs, Annotator~A gives $10$ decisive
``different strategy'' judgements, $11$ decisive ``same strategy,'' and
$10$ uncertain.  The decisive differentiation rate is therefore
$10/21{=}47.6\%$.  This partial separation is expected: SliceGraph
families are defined by co-visitation of non-trivial blocks (a
sub-token-level structural signal), which captures finer process
differences than a human annotator's coarse ``same strategy?'' rubric.
Under a matched strict rubric, Annotator~B's family-level yes-rate is
within $4$\,pp of Annotator~A's, so the family claim is not a
single-annotator artifact.

\subsection{Qualitative Annotation Cases}
\label{app:qual-cases}

Table~\ref{tab:qual-cases} summarises four representative cases used
for qualitative inspection; full anonymised verbatim excerpts are
included in the code repository as
\texttt{annotation\_cases/}.

\begin{table}[h]
\caption{Qualitative annotation cases.  Confidence is on a 1--5 scale.}
\label{tab:qual-cases}
\centering\footnotesize
\begin{tabular}{@{}llllp{0.38\linewidth}@{}}
\toprule
Case & Task & Judgement & Conf. & Diagnostic point \\
\midrule
A1 & BCC within/across & same vs.\ different state & 4/5 &
Answer-finalisation pair vs.\ quadratic-solving control (AIME24) \\
A2 & Articulation transition & strategy shift & 5/5 &
Divisor enumeration $\to$ prime-factorisation pivot (HMMT25) \\
B1 & Within-family & same strategy & 5/5 &
Both runs set up binary-expansion identity identically (HMMT25) \\
B2 & Across-family & process-isomer pair & 5/5 &
Numerical-first vs.\ theory-first route to same answer (HMMT25) \\
\bottomrule
\end{tabular}
\end{table}

\section{Cross-Model Replications}
\label{app:replications}

\subsection{Cross-Scale Replication on Qwen2.5-Math-72B-Instruct}
\label{app:72b-replication}

We replicate the full canonical pipeline---including the reward field,
multi-core decomposition, family-core specialization, and
family-TV---on Qwen2.5-Math-72B-Instruct
($20{,}352$ sampled runs, $20{,}196$ graph-valid rows; $318$ non-code
problems on AIME24/25, BRUMO25, GPQA, HMMT25) under the identical
configuration
($\texttt{sep\_up}{=}8$, $k{=}6$, mutual-$k$NN, $\sigma{=}0.35$,
$M{=}2{,}600$; reward $\alpha{=}0.65$, $\texttt{core\_q}{=}0.75$).

\paragraph{Process-family heterogeneity.}
Under canonical Louvain, 72B retains multi-family structure but
attenuates on saturated math datasets where many problems are solved
near-deterministically.  On GPQA, where the model is less saturated,
the multi-correct-family rate is $\mathbf{86.4\%}$, essentially
matching the 3-model $85.5\%$ headline.

\paragraph{Reward-field and multi-core structure.}
The multi-core rate drops to $42.9\%$ (vs.\ $67.6\%$ primary), consistent
with 72B solving many problems near-deterministically so that the value
landscape has less diversity.  However, on the $81$ cells that \emph{do}
exhibit multi-core structure, family-core specialization is
\emph{higher} ($0.738$ vs.\ $0.59$), indicating sharper family-core
alignment when structural diversity survives.  Family-conditioned
field-max ratios ($1.57\times$) and mean family-TV ($0.117$) match the
primary pattern ($1.55\times$, $0.135$).

\begin{table}[h]
\caption{72B replication summary.  ``Primary'' is the pooled
3-model value cited in the main text; ``72B'' is the matched
replication on the Qwen2.5-Math-72B-Instruct corpus.}
\label{tab:72b-summary}
\centering\footnotesize
\begin{tabular}{@{}lllc@{}}
\toprule
Main construct & Primary (3-model) & 72B & Conclusion \\
\midrule
multi-core rate            & $67.6\%$ & $42.9\%$ & attenuated by saturation \\
family$\times$core spec.\ (multi-core cells) & $0.59$   & $0.738$  & higher when multi-core \\
family / pooled-global field-max & $1.55\times$ & $1.57\times$ & matched \\
multi-correct-family rate  & $85.5\%$ & $67.3\%$ (mean $3.18$/prob) & attenuated by saturation \\
length-variance ratio      & $0.460$  & $0.484$  & not a length artifact \\
mean family-TV             & $0.135$  & $0.117$  & matched \\
\bottomrule
\end{tabular}
\end{table}

Dataset-level values are included in Table~\ref{tab:model-dataset-full}.
The overall pattern is that scaling from 8B to 72B attenuates
multi-core and multi-family rates (via higher accuracy saturation),
but preserves the qualitative atlas structure: when structural
diversity survives, families specialize on disjoint core components
even more sharply than at 8B.

\subsection{Cross-Scale Replication on Qwen3-32B}
\label{app:32b-replication}

We replicate the full canonical pipeline on Qwen3-32B
($20{,}352$ sampled runs; $318$ non-code problems on AIME24/25,
BRUMO25, GPQA, HMMT25) under the identical configuration.
After degenerate-correctness filtering, $147$ reward-evaluable cells
remain.

\paragraph{Key result.}
Qwen3-32B closely matches the primary 3-model atlas picture and
substantially outperforms the 72B replication on multi-family and
multi-core rates.  The multi-correct-family rate ($89.6\%$) exceeds
the primary headline ($85.5\%$); multi-core rate ($75.5\%$) matches
Llama-8B ($78.4\%$); family-TV ($0.143$) and family/global field-max
($1.60\times$) are numerically matched to the primary ($0.135$,
$1.55\times$).  Unlike 72B, Qwen3-32B does not saturate on the math
benchmarks, so the full multi-route structure is preserved at the 32B
scale.

\begin{table}[h]
\caption{Qwen3-32B replication summary.}
\label{tab:32b-summary}
\centering\footnotesize
\begin{tabular}{@{}lllc@{}}
\toprule
Main construct & Primary (3-model) & 32B & Conclusion \\
\midrule
multi-core rate            & $67.6\%$ & $75.5\%$ & matched / above \\
family$\times$core spec.\ (multi-core cells) & $0.59$   & $0.52$  & comparable \\
family / pooled-global field-max & $1.55\times$ & $1.60\times$ & matched \\
multi-correct-family rate  & $85.5\%$ & $89.6\%$ (mean $4.64$/prob) & matched / above \\
length-variance ratio      & $0.460$  & $0.440$  & not a length artifact \\
mean family-TV             & $0.135$  & $0.143$  & matched \\
\bottomrule
\end{tabular}
\end{table}

\subsection{Cross-Architecture Replication on DeepSeek-R1-Distill-Llama-8B}
\label{app:llama-cross-architecture}

The Llama cross-architecture replication ($16{,}512$ runs across
$258$ problems on AIME24, AIME25, GPQA) is the strongest single piece
of architecture transfer evidence.  The same atlas picture reappears:
multi-core solution landscapes, several correct-only families per
problem, and near-identical family-core specialization.

\begin{table}[h]
\caption{Llama-8B replication summary.  ``Primary'' is the pooled
3-model value cited in the main text; ``Llama'' is the matched
replication on the Llama corpus.}
\label{tab:llama-summary}
\centering\footnotesize
\begin{tabular}{@{}lllc@{}}
\toprule
Main construct & Primary (3-model) & Llama & Conclusion \\
\midrule
multi-core rate            & $67.6\%$ & $78.4\%$ & multi-core dominant \\
family$\times$core spec.\  & $0.59$   & $0.597$  & numerically matched \\
family / pooled-global field-max & $1.55\times$ & $1.46\times$ & matches sharpness \\
multi-correct-family rate  & $85.5\%$ & $78.3\%$ (mean $3.98$/prob) & multi-family dominant \\
length-variance ratio      & $0.460$  & $0.469$  & not a length artifact \\
mean family-TV             & $0.135$  & $0.127$  & matched \\
family-TV $>$ null $p_{95}$ & $80.9\%$ & $78.3\%$ ($n{=}152$) & matched \\
\bottomrule
\end{tabular}
\end{table}

Llama also matches the block-type localisation pattern of the reward
field: its high-value landscape is answer-basin dominated
(Table~\ref{tab:reward-field-block-type}), and its family-level
partition remains process-like rather than length-clustered.
Family-TV dynamics replicate as well: across $152$ Llama cells with
valid typed-state kernels, $78.3\%$ exceed the family-label shuffle
null's $p_{95}$, matching the primary rate of $80.9\%$.
Taken together, the Llama replication shows that the atlas story is
not tied to one architecture within the Qwen family.

\end{document}